\DocumentMetadata{}
\documentclass[sigconf]{acmart}
\AtBeginDocument{%
  \providecommand\BibTeX{{%
    \normalfont B\kern-0.5em{\scshape i\kern-0.25em b}\kern-0.8em\TeX}}}
    
\usepackage{hyperref}
\usepackage{hyperxmp}
\usepackage{subcaption}
\usepackage{algorithm,algorithmic}
\usepackage{enumitem}
\usepackage{placeins}
\usepackage{amsfonts}       
\usepackage{pifont}
\usepackage{soul}
\usepackage{nicefrac}       
\usepackage{microtype}      
\usepackage{xcolor}         
\usepackage{utfsym}
\usepackage{multirow}
\usepackage{float}
\usepackage{xspace}
\usepackage{adjustbox}
\usepackage{booktabs}       
\usepackage{MnSymbol}
\usepackage{lipsum}
\usepackage{colortbl}
\newcommand{\cmark}{\ding{51}}%
\newcommand{\xmark}{\ding{55}}%
\sethlcolor{yellow}

\DeclareMathOperator{\transformer}{Transformer\_Encoder}
\DeclareMathOperator{\linear}{Linear}
\DeclareMathOperator{\similarity}{sim}
\usepackage{xspace}

\newcommand{\model}{\texttt{STEDR}\xspace}

\copyrightyear{2025}
\acmYear{2025}
\setcopyright{rightsretained}
\acmConference[KDD '25] {Proceedings of the 31st ACM SIGKDD Conference on Knowledge Discovery and Data Mining V.1}{August 3--7, 2025}{Toronto, ON, Canada.}
\acmBooktitle{Proceedings of the 31st ACM SIGKDD Conference on Knowledge Discovery and Data Mining V.1 (KDD '25), August 3--7, 2025, Toronto, ON, Canada}
\acmISBN{979-8-4007-1245-6/25/08}
\acmDOI{10.1145/3690624.3709418}
\begin{document}


\title[A Deep Subgrouping Framework for Precision Drug Repurposing \\via Emulating Clinical Trials on Real-world Patient Data]{A Deep Subgrouping Framework for Precision Drug Repurposing via Emulating Clinical Trials on Real-world Patient Data}

\author{Seungyeon Lee}
\email{lee.10029@osu.edu}
\affiliation{%
  \institution{The Ohio State University}
  \state{Ohio}
  \country{USA}
}
\author{Ruoqi Liu}
\email{liu.7324@osu.edu}
\affiliation{%
  \institution{The Ohio State University}
  \state{Ohio}
  \country{USA}
}
\author{Feixiong Cheng}
\email{CHENGF@ccf.org}
\affiliation{%
  \institution{Cleveland Clinic}
  \state{Ohio}
  \country{USA}
}
\author{Ping Zhang}\authornote{Corresponding Author}
\email{zhang.10631@osu.edu}
\affiliation{%
  \institution{The Ohio State University}
  \state{Ohio}
  \country{USA}
}

\renewcommand{\shortauthors}{Lee, et al.}

\begin{abstract}
Drug repurposing identifies new therapeutic uses for existing drugs, reducing the time and costs compared to traditional \textit{de novo} drug discovery. Most existing drug repurposing studies using real-world patient data often treat the entire population as homogeneous, ignoring the heterogeneity of treatment responses across patient subgroups. This approach may overlook promising drugs that benefit specific subgroups but lack notable treatment effects across the entire population, potentially limiting the number of repurposable candidates identified. To address this, we introduce \model, a novel drug repurposing framework that integrates subgroup analysis with treatment effect estimation. Our approach first identifies repurposing candidates by emulating multiple clinical trials on real-world patient data and then characterizes patient subgroups by learning subgroup-specific treatment effects. We deploy \model to Alzheimer's Disease (AD), a condition with few approved drugs and known heterogeneity in treatment responses. We emulate trials for over one thousand medications on a large-scale real-world database covering over 8 million patients, identifying 14 drug candidates with beneficial effects to AD in characterized subgroups. Experiments demonstrate \model's superior capability in identifying repurposing candidates compared to existing approaches. Additionally, our method can characterize clinically relevant patient subgroups associated with important AD-related risk factors, paving the way for precision drug repurposing.

\end{abstract}


\maketitle
\vspace{-5pt}
\section{Introduction}
Drug repurposing, the process of identifying new therapeutic uses for existing drugs, has emerged as a promising strategy to accelerate drug development and reduce costs compared to traditional \textit{de novo} drug discovery methods \citep{liu2021deep}. Recent advancements in computational methods have significantly enhanced drug repurposing efforts. These methods utilize various data types, including structural features of compounds or proteins \citep{luo2016dpdr}, genome-wide association studies (GWAS) \citep{sanseau2012use}, and gene expression data \citep{sirota2011discovery}. However, a significant challenge persists in translating pre-clinical outcomes to actual clinical therapeutic effects in humans \citep{buchan2011role}.

\begin{figure}[t]
    \vspace{5pt}
    \centering
    \includegraphics[width=0.4\textwidth]{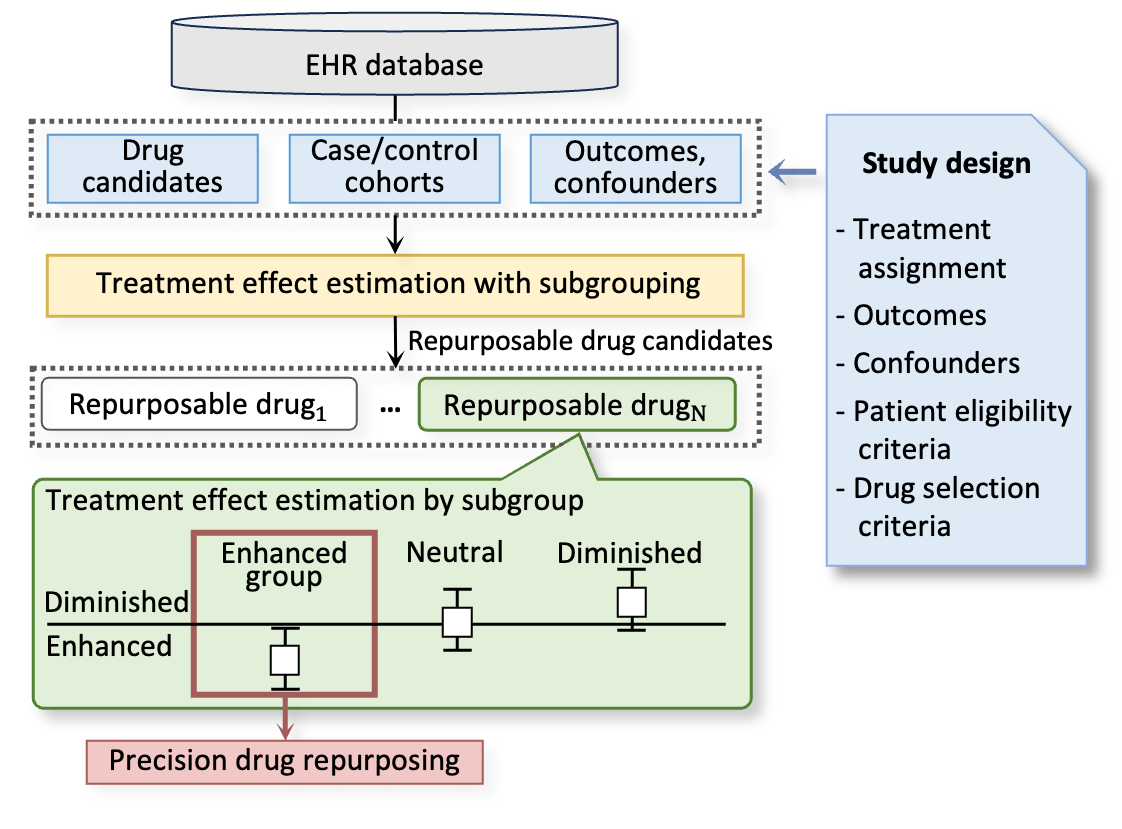}
    \caption{Flowchart of a deep subgrouping framework for precision drug repurposing via emulating clinical trials on real-world data. First, eligible drugs and cohorts are extracted from an EHR database. Second, STEDR estimates the subgroup-specific treatment effects of each drug. Third, repurposable drugs with beneficial effects in patient subgroups will be identified.}\label{fig-0}
\end{figure}

Real-world data (RWD), such as medical claims and electronic health records (EHRs), contains valuable information about patient health outcomes, treatment patterns, and patient characteristics. This makes RWD a vital resource for comparing the treatment effects of drugs and identifying those with beneficial effects as repurposing drug candidates \cite{cheng2024artificial, tan2023drug}. Despite this, existing studies \cite{liu2021deep,zang2022high} often treat the entire population as homogeneous, overlooking the heterogeneity of treatment responses across patient subgroups. This approach may miss promising drugs that benefit specific subgroups but lack notable treatment effects across the entire population, thereby limiting the identification of repurposable candidates.

To address this issue, we aim to identify precision drug repurposing candidates specific to subgroups and to characterize the corresponding patient subgroups. This requires a profound understanding of subgroup-specific treatment effects and the stratification of the entire patient population into distinct subgroups based on their heterogeneous treatment responses. Recently, several methods ~\citep{shi2019adapting, curth2021inductive, schwab2020learning, shalit2017estimating} have been proposed for estimating treatment effects using observational data. Whereas, these methods are not directly applicable to our problem because they are primarily designed for static and low-dimensional data, while RWD is temporal and high-dimensional in nature. In addition, they often fail to characterize heterogeneous subgroups using clinically relevant variables.

In this paper, we introduce Subgroup-based Treatment Effect Estimation for Drug Repurposing (\model), a novel drug repurposing framework that integrates subgroup analysis with treatment effect estimation (TEE). \model aims to estimate the causal effects of treatments on outcomes of interest, given observational covariates while characterizing patient subgroups with heterogeneous treatment effects. We design dual-level attentions (i.e., covariate-level and visit-level) to encode temporal and high-dimensional data into patient representations. Additionally, we develop a subgrouping network that identifies subgroups within the population by learning unique local distributions based on a variational auto-encoder (VAE). This architecture overcomes the challenges of existing treatment effect estimation models that use a single shared latent space for the entire population and fail to differentiate subgroups. We deploy \model to Alzheimer's Disease (AD), a condition with few approved drugs and known heterogeneity in treatment responses. We conduct large-scale drug screening of repurposing drugs for AD treatment through high-throughput clinical trial emulation on MarketScan\footnote{\url{https://www.merative.com/real-world-evidence}} MDCR database — a real-world data representing health services of retirees (aged 65 or older)— which includes over 8 million patients. From a total of 1,134 drugs, \model filters out eligible trial drugs and emulates 100 trials of each drug to estimate their effects. Our experiments show that \model outperforms existing approaches in identifying repurposing candidates and can characterize clinically relevant patient subgroups associated with important AD-related risk factors for precision drug repurposing.
 
In summary, our contributions include:
\begin{itemize}
    \item We propose \model\footnote{\url{https://github.com/yeon-lab/STEDR}}, a novel framework that combines subgroup identification with treatment effect estimation (TEE), aimed at both advancing estimation and facilitating precision drug repurposing. To our knowledge, this is the first study to integrate subgroup analysis into drug repurposing. 
    \item We design dual-level attentions, comprising covariate-level and visit-level components, to investigate the impact of covariates and visits and to learn an individualized patient-level representation.
    \item We develop a subgrouping method that identifies heterogeneous subgroups by learning the unique local distributions that define subgroups based on VAE. 
    \item A real-world deployment on Alzheimer's Disease (AD) demonstrates the effectiveness of \model in identifying potential repurposable candidates and characterizing clinically relevant patient subgroups, demonstrating its capability for precision drug repurposing.
\end{itemize}

\section{Related work}\label{sec-5}
\textbf{Drug repurposing on EHRs.}
EHRs have emerged as a promising resource for drug repurposing as being direct observations from patients,  \cite{tan2023drug}. Several works leverage EHRs for drug repurposing. These methods rely on the treatment effects estimated from the entire population to identify repurposable drug candidates. Specifically, Zang et al. \cite{zang2022high} propose a framework that leverages high-throughput emulations for AD drug repurposing. The framework reweights individuals using the stabilized IPTW derived from a regularized logistic regression-based PS model, and it estimates the treatment effects by adjusted 2-year survival difference and HR. Similarly, studies \cite{xu2023comparing, liu2021deep} employ PS models for IPTW but estimate the treatment effects by directly comparing the outcomes of treated and control groups. Yan et al. \cite{wei2023leveraging} leverage ChatGPT to recommend drugs by expediting the literature review process and evaluating the potential effects of using HR with PS-Matching.

\vspace{3pt}
\noindent \textbf{Treatment Effect Estimation.} Many studies have leveraged the power of neural networks for estimating treatment effects ~\citep{shi2019adapting, curth2021inductive, schwab2020learning, shalit2017estimating}. To apply neural networks for causal inference, several previous works employ a strategy where covariates from different treatment groups are assigned to separate branches. For example, DrangonNet \citep{shi2019adapting} consists of a shared feature network and three auxiliary networks that predict propensity score, and treated and control outcomes, respectively. DR-CRF \citep{Hassanpour2020Learning} learns disentangled representations of the covariates to separately predict the potential outcomes and the propensity score. TransTEE \citep{zhang2022exploring} leverages the Transformer to model the interaction between the input covariate and treatment.

\section{Preliminary}
\noindent
\textbf{Longitudinal Patient Data.}
A patient health record is represented as a sequence of multiple visits in the order of their occurrence, denoted as $\Tilde{\mathbf{x}} = \{\mathbf{x}_1, \cdots, \mathbf{x}_T \}$. Each visit is characterized by a series of varying numbers of diagnosis codes, $m_1, \cdots, m_{|\mathcal{M}|} \in \mathcal{M}$, where $|\mathcal{M}|$ is the number of unique diagnosis codes. The $t$-th visit of the $i$-th patient is expressed as a binary vector $\textbf{x}_{i,t} \in \{0,1\}^{|\mathcal{M}|}$, where a value of 1 for the $m$-th coordinate (i.e. $x_{i,t,m}=1$) indicates that the $m$-th diagnosis code is recorded at that patient's visit. The data is presented as $D \equiv (\Tilde{\mathbf{x}}_i, t_i, y_i)^{N}_{i=1}$, where $\Tilde{\mathbf{x}}_i$ is pre-treatment covariates, $N$ is the number of observed samples. $t_i \in \{0, 1\}$ indicates a treatment variable when binary treatment setting, and $y_i \in \mathbb{R}$ is the observed outcome.

\begin{figure*}[!ht]
    \centering
    \includegraphics[width=\textwidth]{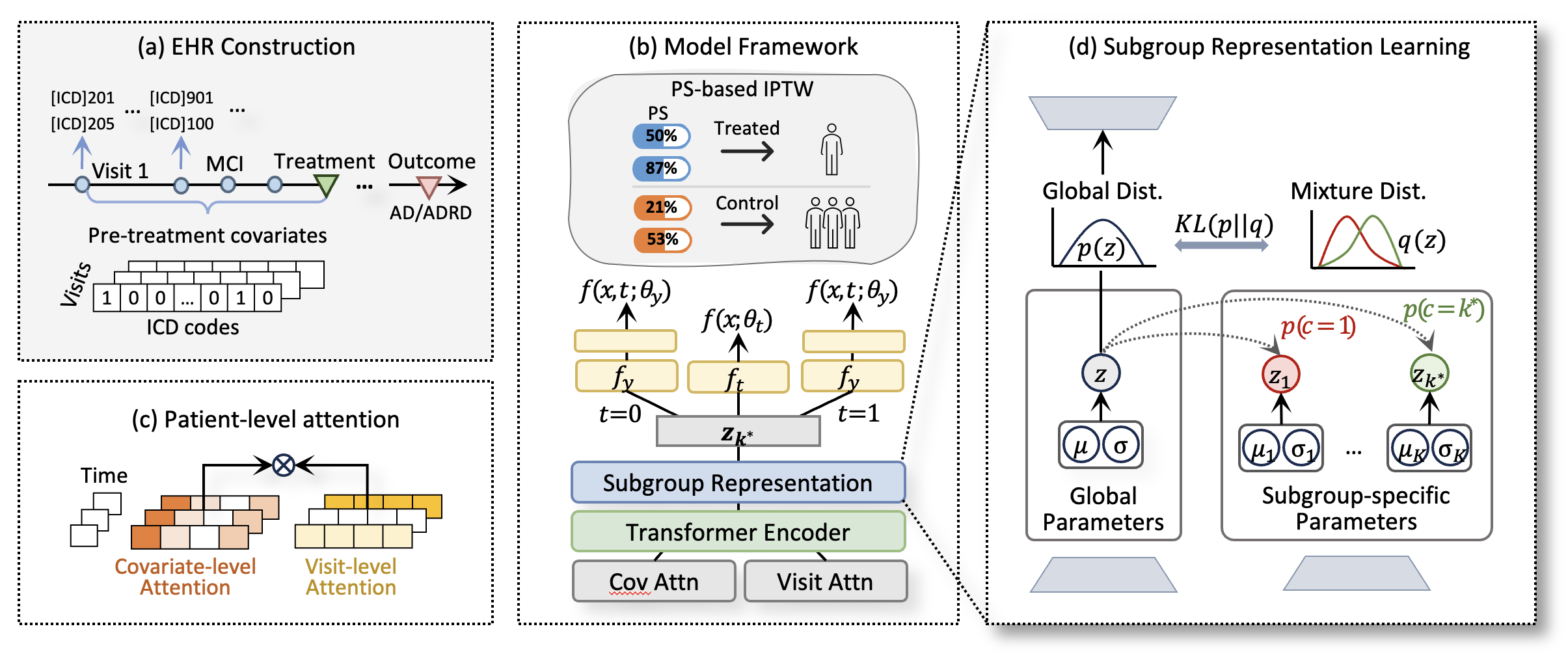}
    \vspace{-15pt}
    \caption{An illustration of \model. The EHR data is processed through patient-level attention to learn individualized representations. The subgroup representation network assigns each subject to a subgroup and extracts subgroup-specific representations. The TEE model predicts the potential outcomes and propensity score from these subgroup-specific representations. The model is trained using the IPTW-based loss for confounder adjustment.}
    \label{fig-model}
\vspace{-7pt}
\end{figure*}

\noindent
\textbf{Treatment Effect Estimation and Subgrouping}
Given a dataset $D \equiv (\mathbf{x}_i, t_i, y_i)^{N}_{i=1}$, a unit $\mathbf{x}$ has two potential outcomes, $Y_{1}(\mathbf{x})$ given it is treated and $Y_{0}(\mathbf{x})$ otherwise, following the potential outcome framework \citep{rubin1974estimating}. The individual treatment effect (ITE) is expressed as $\tau \left (x\right ):=\mathbb{E}[Y|T=1,X]-\mathbb{E}[Y|T=0,X]$. Given an hypothesis $f: \mathcal{X} \rightarrow \mathcal{Y}$, such that $f(\mathbf{x},t) = h_t(\Phi(\mathbf{x}))$, we aim to estimate the treatment effect of the hypothesis $f$ for unit $\mathbf{x}$ as $\hat{\tau}(\mathbf{x})= f(\mathbf{x},1)-f(\mathbf{x},0)$. The treatment effect within a subgroup $k$ can be expressed as: $\tau_{k} = \mathbb{E}[Y|T=1, X \in k]-\mathbb{E}[Y|T=0, X \in k]$. The objective is to identify subgroups of the population $\{C_1,...C_K\}$, such that $\bigcup_{i=1}^K C_i = D$ and $C_i \cap C_j= \phi$ for $i\neq j$. Covariates in each subgroup $C_k$ follows a specific distribution, $\mathcal{N}(x|\mathbf{\mu}_k, \Sigma_k)$, and the treatment effect is estimated within each subgroup, such that $\hat{\tau}_k = f(\mathbf{x},1)-f(\mathbf{x},0)$ for $\mathbf{x}\in C_k$. The output is the treatment effect estimation $\hat{\tau}$ and subgroups $\{C_1,...C_K\}$. 

Our framework first estimates the treatment effect, $\hat{\tau}(\mathbf{x})$, while identifying subgroups with heterogeneous effects. \model then identifies repurposable drug candidates with enhanced $\hat{\tau}(\mathbf{x})$ across all subgroups. The causal assumptions underlying our study are detailed in Section \ref{sec-ci} of the Supplementary material.


\section{Method}\label{sec-2}
TEE allows for a quantification of the potential benefits or risks of the drug, which enables the accurate assessment of the existing drugs' effects in new therapeutic uses. We leverage TEE as a foundational component of our framework for drug repurposing. Our method, \model, innovatively combines TEE with subgroup identification to estimate subgroup-specific treatment effects for targeted drug repurposing. Figure \ref{fig-model} shows an illustration of \model. 

\vspace{-5pt}
\subsection{Patient-level Representation}
\textbf{Covariate-level Attention.} We investigate the impact of each covariate by covariate-level attention. We assume that each subgroup has distinct characteristics, so the attentions affect both subgroup assignment and the potential outcome. 

To preserve differences between covariates from being lost, the method learns a different embedding for each covariate. Each covariate is expressed as a sequence of its occurrences across all visits. This sequence is denoted as $\boldsymbol{d}_{i,m} = \{x_{i,1,m},\cdots,x_{i,T,m}\}$ for the $m$-th code of the $i$-th patient. This vector not only reflects how often the code appears but also preserves the order of its occurrences. The embedding for each covariate is denoted as $\boldsymbol{h}_{i,m} = \textrm{emb}^m_d (\boldsymbol{d}_{i,m})$, where $\boldsymbol{h}_{i,m} \in \mathbb{R}^{p}$. 

Patient data poses unique challenges due to irregular temporality, meaning variation in time intervals between visits. Since these intervals often play a crucial role in the outcome, we also incorporate temporal information in covariate embeddings. The time information is measured as the relative time to the time of interest for each visit and is expressed as $\boldsymbol{t}_{i} = {t_1,\cdots,t_T}$. The embedding for time information is expressed as $\boldsymbol{r}_{i} = \textrm{emb}_t(\boldsymbol{t}_{i})$, where $\boldsymbol{r}_{i} \in \mathbb{R}^{p}$. The final covariate embedding for the $m$-th code is then obtained as follows:
\begin{equation}\label{eq-1}
    \boldsymbol{e}_{m} = \boldsymbol{h}_m + \boldsymbol{r}
\end{equation}
Note that $\boldsymbol{r}_{i}$ is applied to all covariates for the $i$-th patient. To learn and signify the importance of each covariate, we introduce a learnable vector $\boldsymbol{s}^d$. The attention scores are computed as $\boldsymbol{a}^d \in \mathbb{R}^{|\mathcal{M}|\times 1}$, where:
\begin{equation}\label{eq-2}
    a^{d}_{m} = \frac{\exp(\boldsymbol{e}_m^T\boldsymbol{s}^d)}{\sum_m^{\mathcal{M}}\exp(\boldsymbol{e}_m^T\boldsymbol{s}^d)}
\end{equation}

\noindent
\textbf{Visit-level Attention.} Each patient's data is represented as a series of all visits. We also examine the importance of each visit. The embedding for each visit is obtained as $\boldsymbol{v}_{i,t}= \textrm{emb}_v(\textbf{x}_{i,t})$. Note that visit embeddings for all visits are derived from a single embedding layer. We also introduce a learnable vector $\boldsymbol{s}^v$ and calculate attention scores $\boldsymbol{a}^v \in \mathbb{R}^{T \times 1}$ as follows:
\begin{equation}\label{eq-3}
    a^{v}_{t} = \frac{\exp(\boldsymbol{v}_t^T\boldsymbol{s}^v)}{\sum_t^T\exp(\boldsymbol{v}_t^T\boldsymbol{s}^v)}
\end{equation}

\noindent
\textbf{Patient-level Attention.} With covariate-level and visit-level attention scores, our method obtains patient-level importance, computed as $\boldsymbol{A} = \boldsymbol{a}_v \boldsymbol{a}_d^T \in \mathbb{R}^{T \times |\mathcal{M}|}$. The input covariates $\Tilde{\mathbf{x}}$ are element-wise multiplied with $\boldsymbol{A}$, denoted as $\boldsymbol{A}\odot \Tilde{\mathbf{x}}$. This enables the model to make more patient-specific predictions and decisions, focusing on the most relevant information from covariates and visits. $ \Tilde{\mathbf{x}}$ is then input into the one-layer feedforward neural networks with ReLU and the Transformer encoder \citep{vaswani2017attention} to extract the latent features in sequential context as:
\begin{equation}\label{eq-4}
    \hat{\textbf{x}} = \transformer(\linear(\boldsymbol{A}\odot \Tilde{\mathbf{x}}))
\end{equation}

\subsection{Subgroup Representation Learning}
Subgroup representation learning aims to identify potential subgroups by capturing the heterogeneous data distributions and learning subgroup-specific representations. We rely on the following assumptions. First, we assume that heterogeneous subgroups exist within the data, where each subgroup follows a distinct local distribution that captures the specific attributes of that subgroup. The entire population follows a single global distribution reflecting the characteristics of the entire population. Second, the global distribution can be estimated as a mixture Gaussian of the local distributions.

The entire population follows the global distribution, with $p$-dimensional global parameter vectors $\{\mu_g, \sigma_g\}$.
\begin{equation}\label{eq-5}
p(\mathbf{z}) \sim \mathcal{N}(\mathbf{z}|\mathbf{\mu}_g,\Sigma_g)
\end{equation}
We infer $p(\mathbf{z})$ using $\Phi(\hat{\mathbf{x}}_i; \phi_g)$, which is the global encoder of the VAE that extracts global parameters. The global encoder $\Phi(\cdot;\phi_g)$ and decoder $g_{\phi}$ of the VAE are optimized using the reconstruction loss to ensure that the global distribution accurately reflects the latent characteristics of the entire population.
\begin{align}\label{eq-12}
    \mathcal{L}_{vae} & =  \int p(\mathbf{z}|\hat{\mathbf{x}}) \log g_{\phi}(\mathbf{x}|\mathbf{z}) d\mathbf{z} =  \int \Phi(\hat{\mathbf{x}}_i;\phi_g) \log g_{\phi}(\mathbf{x}|\mathbf{z}) d\mathbf{z}
\end{align}

The $k$-th subgroup's local distribution is also characterized by $p$-dimensional local parameter vectors $\mu_k$ and $\sigma_k$ as follows:
\begin{equation}\label{eq-6}
q_k(\mathbf{z}) \sim \mathcal{N}(\mathbf{z}|\mathbf{\mu}_k,\Sigma_k)
\end{equation}
We infer $q_k(\mathbf{z})$ using $\Phi(\hat{\mathbf{x}}_i; \phi_{k})$, which is the $k$-th local encoder. Note that the global parameters and the local parameters are distinctly extracted from different one-layer encoders. 

Given a sample $\hat{\mathbf{x}}$, the global distribution can be estimated as a mixture Gaussian distribution of the local distributions with probabilities of the sample assigned to the subgroups:
\begin{equation}\label{eq-8}
p(\mathbf{z}) \sim q(\mathbf{z}):= \sum_k^K q_k(\mathbf{z})\cdot p(c=k|\hat{\mathbf{x}}) 
\end{equation}
The probability assigned to the subgroup $k$, $p(c=k|\hat{\mathbf{x}}_i)$, can be rewritten, derived from the Bayes’ theorem: 
\begin{align}\label{eq-7}
    p(c=k|\hat{\mathbf{x}}_i) & = \frac{p(\hat{\mathbf{x}}_i|c=k)\cdot p(c=k)}{p(\hat{\mathbf{x}})} \\ \notag
    & =\frac{p(\hat{\mathbf{x}}_i|c=k)\cdot p(c=k)}{\sum_{k'}^{K} p(\hat{\mathbf{x}}_i|c=k')\cdot p(c=k')}
\end{align}
$p(\hat{\mathbf{x}}_i|c=k)$ can be estimated using the global distribution. Consequently, we estimate the subgroup probability as the softmax function over similarities between the global and local representations of all subgroups:
\begin{equation}\label{eq-7}
\vspace{-3pt}
p(c=k|\hat{\mathbf{x}}_i)=\frac{\exp{(\similarity{(\Phi(\hat{\mathbf{x}}_i;\phi_k),\Phi(\hat{\mathbf{x}}_i;\phi_g))})}}{\sum_{k'}^K \exp{ (\similarity{( \Phi( \hat{ \mathbf{x}}_i; \phi_{k'}), \Phi(\hat{\mathbf{x}}_i;\phi_g)})}}
\end{equation}%
\noindent where $\similarity(\cdot)$ indicates the similarity score, which is calculated using the Euclidean distance.

\begin{algorithm}[t]
\caption{Training Process of STEDR}
\label{alg:algorithm}
\begin{flushleft}
\textbf{Input}: Dataset $\mathcal{D}$, The number of subgroups $K$\\
\textbf{Parameter}: The patient representation network, the subgroup representation network, the prediction network\\
\textbf{Output}: estimated treatment effect $\hat{\tau}$, subgroups $\{C_i, ...,C_K\}$ 
\end{flushleft}
\begin{algorithmic}[1] 
    \STATE Initialize network parameters
    \REPEAT
    \STATE Obtain patient level representations $\hat{\textbf{x}}$ by Eqs. (\ref{eq-2})-(\ref{eq-4})
    \STATE Obtain the local representation of each subgroup by Eq. (\ref{eq-6})
    \STATE Compute subgroup probability  $p(c=k|\hat{\mathbf{x}})$ by Eq. (\ref{eq-7})
    \STATE Assign the subgroup, such that $k^*=\textrm{argmax}_k p(c=k|\hat{\mathbf{x}})$
    \STATE Predict outcomes, propensity scores, and assigned subgroups in the prediction network $f(\cdot;\Theta)$
    \STATE Update network parameters by Eq. (\ref{eq-total})
    \UNTIL convergence
\end{algorithmic}
\vspace{-3pt}
\end{algorithm}

The local distributions are learned by KL Divergence between the mixture Gaussian distribution and the global distribution. It ensures that the local representations are finely tuned to reflect the unique attributes of each subgroup, and the global distribution reflects the diverse characteristics of all subgroups.
\begin{equation}\label{eq-9}
\vspace{-3pt}
    \mathcal{L}_{kl} = \sum_z p(\mathbf{z})\log{\left (\frac{p(\mathbf{\mathbf{z}})}{q(\mathbf{z})}\right )}
\end{equation}

To balance the local distributions, we also define a target distribution as a reference, following \citep{xie2016unsupervised}. This target distribution is defined using soft assignments to subgroups and is used to guide the learning process, expressed as:
\begin{equation}\label{eq-10}
\vspace{-3pt}
q(c=k|\hat{\mathbf{x}}_i)=\frac{p(c=k|\hat{\mathbf{x}}_i)^2/r_k}{\sum_{k'}^{K} p(c=k'|\hat{\mathbf{x}}_i)^2/r_{k'}} ,\quad r_k=\sum_{i}p(c=k|\hat{\mathbf{x}}_i) 
\end{equation}
\noindent where $r_k$ represents the soft assignment frequencies to subgroup $k$, and $p(c=k|\hat{\mathbf{x}}_i)$ calculates the soft assignment probability for subgroup $k$ given $\hat{\mathbf{x}}_i$. The overall loss function, $\mathcal{L}_{target}$, is defined as follows:
\begin{equation}\label{eq-11}
\vspace{-3pt}
    \mathcal{L}_{td}=\sum_i\sum_k q(c=k|\hat{\mathbf{x}}_i)\cdot\log\frac{q(c=k|\hat{\mathbf{x}}_i)}{p(c=k|\hat{\mathbf{x}}_i)}
\vspace{-3pt}
\end{equation}

The total loss function for the subgroup representation network is as follows:
\begin{equation}\label{eq-13}
\vspace{-3pt}
    \mathcal{L}_{snn} =  \mathcal{L}_{kl} + \mathcal{L}_{td} + \mathcal{L}_{vae}
\end{equation}

Each sample is assigned to the subgroup with the highest probability, such that $k^*=\textrm{argmax}_k p(c=k|\hat{\mathbf{x}})$. The representation extracted from the corresponding subgroup's encoder, $\Phi(\cdot;\phi_{k^*})$, is fed into the prediction network.

\subsection{Treatment Effect Estimation for Drug Repurposing}

\noindent\textbf{Outcome Prediction.}
Given the representation of the assigned subgroup, the prediction network estimates both treated and control outcomes. To preserve treatment information within the high-dimensional latent representation, the network assigns features from distinct treatment groups to separate branches. Furthermore, we design additional modules to predict treatment assignment to balance the distributions of treated and control groups and to predict the subgroup assignment to ensure that the latent representations effectively distinguish the unique characteristics of subgroups. 

The prediction network $f(\cdot;\Theta)$ is composed of four separate feedforward networks: $\{f(x,t; \theta_y): \mathbb{R}^p\rightarrow\mathbb{R}\}$ for predicting potential outcomes $\hat{y}$ (two heads for $t\in\{0,1\}$); $f(x; \theta_t): \mathbb{R}^p\rightarrow\mathbb{R}$ for the treatment assignment $\hat{t}$; $f(x; \theta_k): \mathbb{R}^p\rightarrow\mathbb{R}^K$ for subgroup assignment $\hat{k}$. The prediction network is optimized as follows:
\begin{align}\label{eq-14}
\vspace{-5pt}
\mathcal{L}_{pnn} = \sum_{i=1}^{|D|} &
\mathrm{CE}(f(\Phi_{k^*_i}(\hat{\mathbf{x}}_i);\theta_t),t_i) + \mathrm{CE}(f(\Phi_{k^*_i}(\hat{\mathbf{x}}_i);\theta_k),k^*_i) \notag \\ 
& + w_i\cdot \ell(f(\Phi_{k^*_i}(\hat{\mathbf{x}}_i),t_i;\theta_y),y_i)
\vspace{-3pt}
\end{align}
\noindent where $\mathrm{CE}(\cdot)$ is the cross-entropy loss and $\ell(\cdot)$ represents the mean squared error for continuous outcomes or the cross-entropy loss for binary outcomes. The term $w_i$ refers to individual weights.

\vspace{5pt}
\noindent\textbf{Propensity Score Weighting.}
The weights $w$ aim to reweight the population for confounder adjustment, ensuring that the treated and control groups are comparable and mitigating the influence of confounding variables. They are computed using inverse probability of treatment weighting (IPTW), with the predicted probability of receiving treatment $\hat{t}_i$ and the probability of being in the treated group $Pr(T)$, expressed as $w_{i} = Pr(T)/\hat{t}_{i} + (1-Pr(T))/(1-\hat{t}_{i})$.

\vspace{5pt}
\noindent\textbf{Loss Function and Optimization.}
To identify subgroups with different treatment effects, we introduce an additional loss function aimed at reducing the overlap between confidence intervals (CIs) of estimated treatment effects across subgroups, which forces the distinct separation of subgroups. The overlap for two intervals is defined as $OL(\text{CI}_i, \text{CI}_j) = \max(0, \min(\text{up}_i, \text{up}_j) - 
    \max(\text{low}_i, \text{low}_j))$, where $\text{CI}_i$, $\text{up}_i$, and $\text{low}_j$ represent the CI, upper bound, and lower bound of subgroup $i$, respectively.
    
The total overlap penalty given a batch is computed by summing these overlaps for every pair of subgroups, with a strength factor $\alpha$, which is expressed as:
\begin{align}\label{eq-17}
\vspace{-3pt}
    \mathcal{L}_{overlap} = \alpha \cdot \sum_{i=1}^{K}\sum_{j=i+1}^{K} OL(\text{CI}_i, \text{CI}_j)
\vspace{-3pt}
\end{align}

The total loss function to optimize the model is expressed as Eq. (\ref{eq-total}). The training process of the model is outlined in Algorithm \ref{alg:algorithm}.
\begin{equation}\label{eq-total}
\vspace{-5pt}
    \mathcal{L} = \mathcal{L}_{snn} + \mathcal{L}_{pnn} + \mathcal{L}_{overlap}
\end{equation}

\vspace{3pt}
\noindent
\textbf{Drug Repurposing with Patient Subgroups.}
Given an eligible patient cohort for each drug, our framework first estimates the subgroup-specific treatment effects. We then calculate the 95\% confidence intervals (CIs) and evaluate the statistical significance of the averages of estimated treatment effects for each subgroup over 100 trials. The framework identifies repurposable drugs that have enhanced effects with $p<0.05$ across all or specific subgroups. 


\section{Deployment of Drug Repurposing for Alzheimer's Disease}\label{sec-4}

This study conducts large-scale drug screening of repurposing drugs for Alzheimer’s disease (AD) treatment via high-throughput clinical trial emulation on RWD containing over 8 million patients. AD is a highly heterogeneous neurodegenerative disorder, and drug effects may vary based on genetic risk factors and subtypes \citep{paranjpe2019insights}. In this study, we evaluate the subgroup-based heterogeneous effects of trial drugs on the progression of patients with mild cognitive impairment (MCI) to AD and AD-related dementias (ADRD) \citep{xu2023comparing, duan2020leverage, zang2023high}. From a total of 1,134 drugs, \model emulates 100 trials of each eligible drug to identify new repurposable drug candidates for AD. The following sections detail our data and study design.

\vspace{-5pt}
\subsection{Data}
We used a large-scale real-world longitudinal patient-level healthcare warehouse, MarketScan\footnote{\url{https://www.merative.com/real-world-evidence}} Medicare Supplemental and Coordination of Benefits Database (MDCR) from 2012 to 2018, which is a claims database that represents health records for over 8 million retirees (aged 65 or older) in the USA. The MarketScan data contains individual-level and de-identified healthcare claims information, including diagnoses, procedures, prescriptions, and demographic characteristics. We identify 155K distinct MCI patients, among whom 40K are diagnosed with AD. MCI and AD patients are identified using the diagnosis codes. The diagnosis codes are defined by the International Classification of Diseases (ICD) 9/10 codes. We map the ICD codes to Clinical Classifications Software (CCS), including a total of 286 codes. For drugs, we match national drug codes (NDCs) to ingredient levels. We use diagnosis codes and their time information to construct input variables for our method.

\begin{figure}[!t]
\centering
\includegraphics[width=0.5\textwidth]{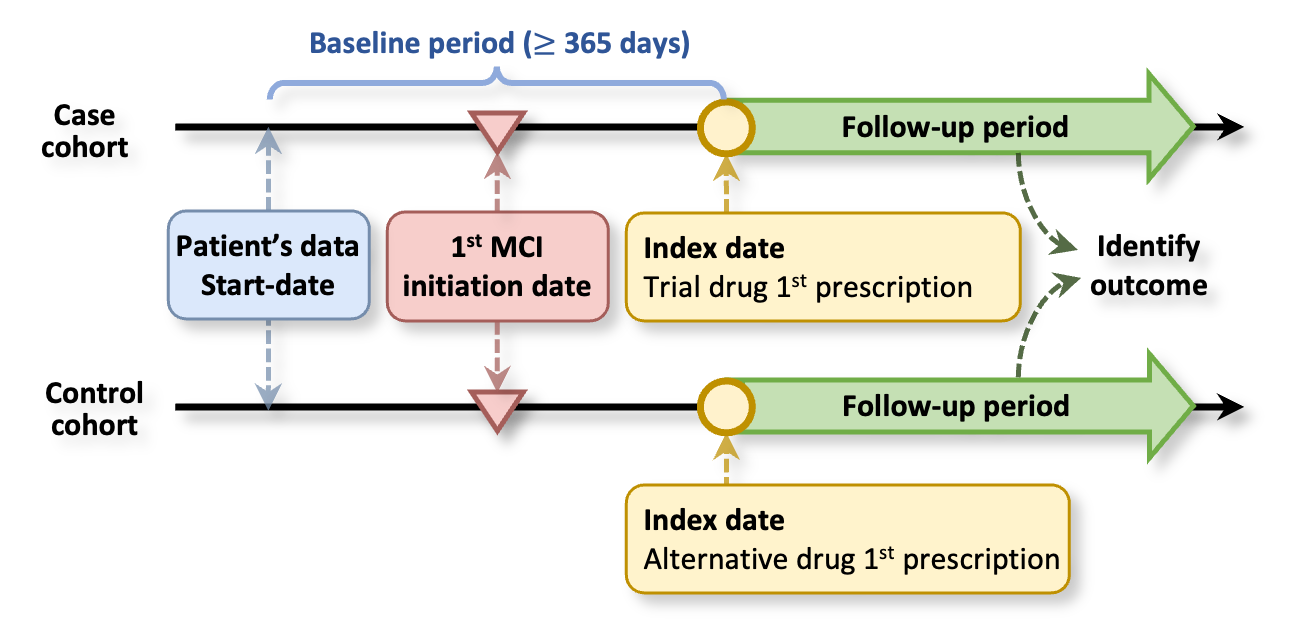}
\vspace{-15pt}
\caption{Study design. The baseline and follow-up periods encompass all dates before and after the index date, respectively. Patients in the study cohort: (1) were diagnosed with MCI before the index date; (2) had no history of AD or ADRD before the index date; (3) had at least one year of medical records before the index date.}
\label{fig-sd}
\vspace{-5pt}
\end{figure}

\begin{figure}[!t]
\centering
\includegraphics[width=0.5\textwidth]{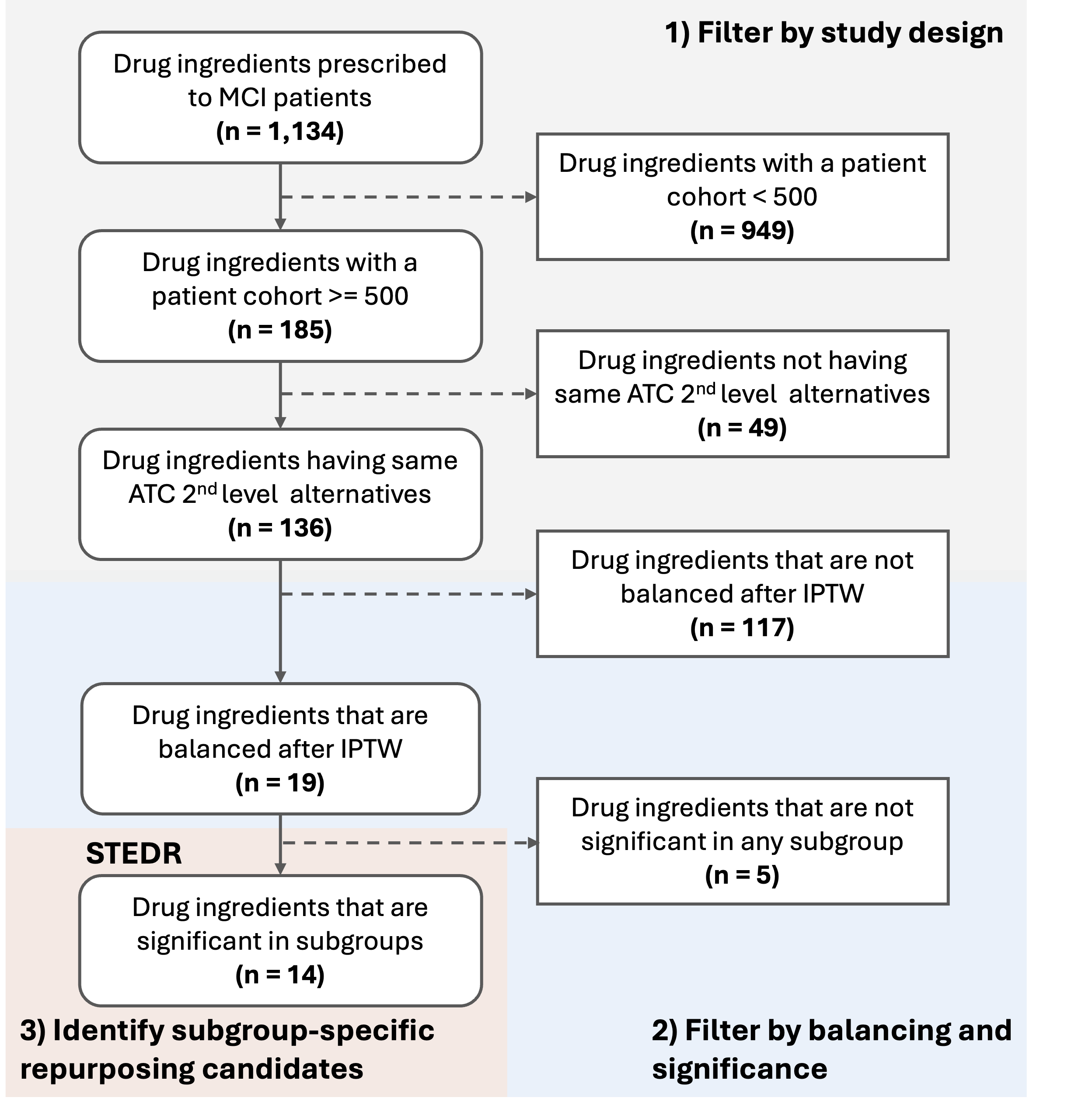}
\vspace{-10pt}
\caption{Drug selection and screening criteria. From 1,134 drugs in the dataset, \model filters out 19 trial drugs that met our study design and balancing criteria. Among them, \model identifies 14 drug candidates that are significant in subgroups.} 
\label{fig-screening}
\vspace{-10pt}
\end{figure}

\vspace{-5pt}
\subsection{Study Design}
\noindent\textbf{Case and Control Cohorts.} 
The estimation of drug effects necessitates a comparative analysis between two cohorts: the case cohort with patients prescribed a trial drug, and the control cohort prescribed alternative drugs. Figure \ref{fig-sd} shows our study design with patient inclusion criteria. Patients are excluded if they had a history of AD/ADRD before the index date or had an age < 50 at the time of the MCI diagnosis. To prevent overlap between the case and control cohorts, patients in the control cohort are excluded if they have been prescribed the trial drug. All patients are followed up to 2 years, until they are diagnosed with AD, or loss to follow-up, whichever occurs first. 

\noindent\textbf{Covariates and Outcomes.} The baseline period is leveraged to construct the pre-treatment covariates, while the follow-up period is used to evaluate the drug effect: patients diagnosed with AD/ADRD during this period are labeled as positive ($y=1$), otherwise negative ($y=0$). A negative value of the TE indicates an enhanced effect of the drug, potentially effective in reducing the incidence of AD/ADRD. 


\vspace{5pt}
\noindent\textbf{Target Trial Emulations.} We emulate high-throughput trials to evaluate the effect of each trial drug, with various control cohorts: (1) prescribed a random alternative drug, and (2) prescribed a similar drug classified under the same ATC-L2 (second-level Anatomical Therapeutic Chemical classification codes) as the trial drug \cite{zang2022high}. We emulate 100 trials for each drug, of which 50 trials are random controls and the others are ATC-L2 controls. The number of patients in the control cohort is set to a maximum of 3 times the case cohort. 

\vspace{5pt}
\noindent\textbf{Drug Selection Criteria.}
We conduct large-scale drug screening of repurposing drugs for AD using RWD via a high-throughput clinical trial emulation. Given the initial set of all drugs under consideration, \model filters out eligible drugs, as described in Figure \ref{fig-screening}. Balanced trial drugs are evaluated by standardized mean differences (SMD) \cite{austin2009using} and the weighted propensity area under the curve (AUC) using IPTW \cite{shimoni2019evaluation}. \model identifies repurposable drug candidates with enhanced ATE in the population or HTE in specific subgroups, with $p$-values $\leq 0.05$ measured as the difference from zero. We use an adjusted $p$-value \citep{benjamini1995controlling}. 

In our framework, we assume three potential subgroups: enhanced, neutral, and diminished groups. Additional information, including data, definitions of MCI and AD/ADRD, metrics, and implementation details are provided in Section \ref{sec-AD} of the Supplemental material.

\vspace{-5pt}
\section{Results}\label{sec-4}
We demonstrate the performance of our model, focusing on answering the following research questions:
\begin{itemize}[leftmargin=*]
    \item \textbf{Q1: How accurate is \model on TEE?} 
    \item \textbf{Q2: How effective is \model on drug repurposing?}
    \item \textbf{Q3: How does \model enhance precision drug repurposing?}
\end{itemize}

\vspace{-5pt}
\subsection{Q1: How accurate is the method on TEE?}
TEE is an effective approach in drug repurposing, as it quantifies the magnitude of the drug effect. The accuracy and reliability of TEE models, therefore, are directly linked to the success of drug repurposing efforts. We conduct a quantitative analysis of TEE using synthetic and semi-synthetic datasets, comparing the proposed method against state-of-the-art neural network-based TEE models \citep{pmlr-v130-curth21a, shi2019adapting, shalit2017estimating, Hassanpour2020Learning, schwab2020learning, nie2021vcnet, zhang2022exploring}. Additionally, we compare with two representative subgrouping models \citep{lee2020robust, nagpal2020interpretable} to further evaluate the effectiveness of the proposed method in subgroup identification.  

Our method outperforms the baselines on both subgrouping and TEE tasks across all datasets, especially representing a significant reduction in error on TEE performance. Specifically, it achieves 35.3\%, 5.6\%, and 35.1\% lower error than the second-best model, as measured by the precision in the estimation of heterogeneous effect (PEHE). The datasets, baselines, evaluation metrics, experimental setting, and results are detailed in Section \ref{sec-simulation} of the Supplemental material.

\begin{table}[t]
\centering
\caption{Comparison of drug repurposing approaches based on model type, measure (ATE: average treatment effect; HR: hazard ratio; HTE: heterogeneous treatment effect), and consideration of time-varying covariate information (Time) and subgrouping (Group).}\label{tb-compare}
\vspace{-10pt}
\begin{tabular}{llccc}
\toprule
Method & Backbone & Measure & Time & Group \\ \midrule
Liu et al. \cite{liu2021deep} & LSTM & ATE & \cmark & \xmark \\
Xu et al. \cite{xu2023comparing} & LR & ATE & \xmark & \xmark \\
Zhang et al. \cite{zang2022high} & LR & HR & \xmark & \xmark \\
Charpignon et al. \cite{charpignon2022causal} & LR & HR & \xmark & \xmark \\ \midrule
\textbf{\model} (ours) & Transformer & HTE & \cmark & \cmark \\
 \bottomrule
\end{tabular}
\end{table}

\begin{table*}[h]
\caption{Comparison of drug repurposing methods by the number of identified drug candidates. Drugs marked with $^\dagger$ are significant in certain subgroups but not in the overall population, which can be missed by existing work.
}\label{tb-drug-num}
\begin{tabular}{lclcc}
\toprule
 && Identified Drug Repurposing Candidates && Total Number \\ \midrule
PS-IPTW (LSTM) && Bupropion, Gabapentin, Rosuvastatin, Trazodone && 4 \\
PS-IPTW (LR) && Gabapentin, Rosuvastatin, Trazodone && 3 \\ \midrule
\model (ours) && \begin{tabular}[c]{@{}l@{}}Acetaminophen, Amoxicillin, Bupropion, Citalopram$^\dagger$, \\ Fluticasone$^\dagger$, Gabapentin, Lisinopril$^\dagger$, Losartan$^\dagger$, Metformin$^\dagger$, \\ Nystatin$^\dagger$, Pantoprazole$^\dagger$, Pravastatin$^\dagger$, Rosuvastatin, Trazodone\end{tabular}&& 14 \\ \bottomrule
\end{tabular}
\end{table*}

\begin{figure*}[t]
\centering
\includegraphics[width=\textwidth]{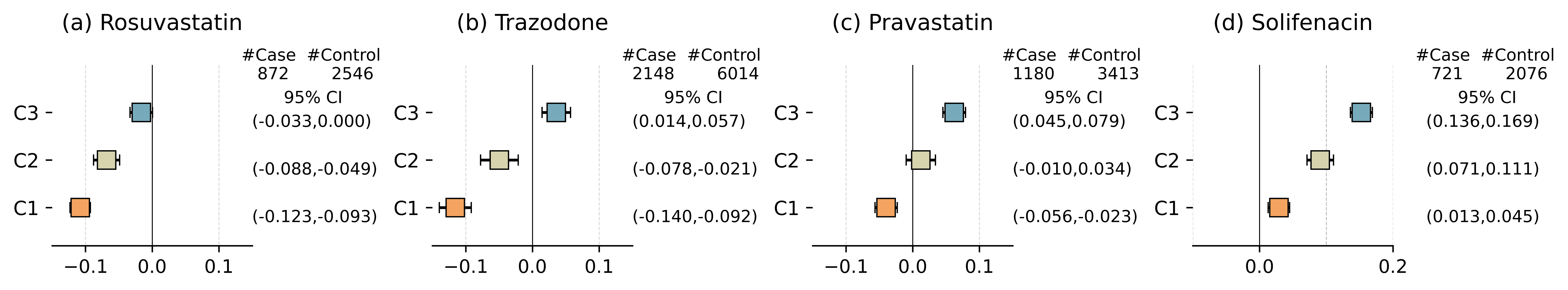}
\vspace{-18pt}
    \caption{Visualization of 95\% confidence intervals of estimated treatment effects across different patient subgroups from 100 trials. C1 to C3 represents Subgroups 1 to 3. We show the results of four drugs, which represent four categories of identified repurposing candidates: (a) significant in all three subgroups, (b) significant in two of three subgroups, (c) significant in one of three subgroups, and (d) not significant in any subgroups. Results of the full list of 14 drugs are presented in Figure \ref{fig-all-drug} and Table \ref{tb-all-drug} of the Supplemental material.}
    \label{fig-mp}
\end{figure*}

\subsection{Q2: How effective is the method on drug repurposing?}
We screened 1,134 drugs (with 100 emulated trials for each drug) and found 136 that met our study design criteria. From the eligible trials of these drugs, 19 were balanced after applying IPTW (Figure \ref{fig-screening}). To identify significant repurposable drug candidates, we first estimated the ATE of these drugs. Among the 19 drugs, 6 drugs showed enhanced effects in the entire population, listed in Table \ref{tb-drug-num} of the Supplemental material without highlights. Additionally, we identified 8 more subgroup-targeted drug candidates with enhanced HTE in specific subgroups, with $p$-values $\leq 0.05$ (highlighted in the table), suggesting these drugs are beneficial to certain subpopulations.

We compare our method with representative drug repurposing approaches based on model type, measure, and consideration of time-to-event information (Time) and subgrouping (Group) in Table \ref{tb-compare}. The ATE-based approaches by \cite{liu2021deep, xu2023comparing} directly estimate the ATE to provide an overall effect of the drug after confounding adjustment via the propensity score (PS)-based IPTW (PS-IPTW) using long short-term memory (LSTM) that incorporates time-varying covariates and logistic regression (LR) networks. The methods by \cite{zang2022high, charpignon2022causal} estimate hazard ratio (HR) after confounding adjustment via PS-IPTW using LR. These approaches, while useful and straightforward, have limitations. They focus on identifying repurposable drugs with beneficial effects across the entire population, potentially overlooking variations in treatment responses without subgrouping. This may result in ignoring the risks of the drug for some subpopulations or missing drugs that may be highly effective for specific subgroups, ultimately leading to missed opportunities for effective treatment strategies. Our method, \model, employs Transformer for TEE and heterogeneous treatment effect (HTE) analysis for subgrouping. We compare the PS-IPTW methods on the number of identified candidates in Table \ref{tb-drug-num}. Our method, \model, identified 14 candidates, with 8 additional drugs that have enhanced effects in specific subpopulations, whereas the PS-IPTW (LSTM) and PS-IPTW (LR) methods identified 4 and 3 drug candidates, respectively.

Our approach addresses limitations in existing approaches by incorporating subgroup-specific treatment effects into clinical decision-making, rather than focusing on the entire population. \model identifies additional potential drugs that are not discovered from existing methods, by accounting for variability of drug effects in the population. This enables us to expand opportunities for drug repurposing by finding potential drugs that meet the specific needs of each subgroup.



\vspace{-5pt}
\subsection{Q3: How does \model enhance precision drug repurposing?}
\vspace{2pt}
\noindent\textbf{Subgroup-Specific Treatment Effect Analysis} Figure \ref{fig-mp} presents the 95\% CIs of the estimated treatment effects in Subgroups 1 to 3, with Subgroup 1 showing the most enhanced effect and Subgroup 3 exhibiting the most diminished effect. We select four example drugs as case studies, which represent a range of response patterns across the subgroups:
\vspace{-3pt}
\begin{itemize}[leftmargin=*]
    \item \textit{Rosuvastatin}: This drug demonstrated enhanced ATE, also with HTE consistently below zero across all subgroups. This suggests that Rosuvastatin could be broadly applicable as a repurposable drug for AD.
    \item \textit{Trazodone}: Although the drug showed beneficial effects in the overall population, some patient subgroups were at risk with these drugs, suggesting that they may not be broadly applicable. This highlights that their application might need to be tailored to avoid risks in certain subgroups, with the importance of identifying subgroup-specific effects.  
    \item \textit{Pravastatin}: The drug exhibited diminished effects in the overall population, but specific subgroups showed potential benefits. This indicates their potential for targeted repurposing. 
    \item \textit{Solifenacin}: This drug showed diminished effects across both the entire population and all subgroups, suggesting limited utility for repurposing in the context of AD. 
\end{itemize}

These observations underscore the complexity and variability in treatment responses in the population, which is consistent with the complex nature of AD. \model effectively addresses this variability by capturing a wide range of responses across potential patient subgroups. The subgroup-targeted treatment strategy advances precision drug repurposing, ensuring that treatment strategies are not just broadly applicable, but finely tuned to the unique characteristics and needs of subgroups, thereby reducing the risk of adverse effects. \model is particularly vital in managing diseases like AD/ADRD, where responses can vary greatly among patients. Note that the results for all trial drugs are presented in Figure \ref{fig-all-drug} and Table \ref{tb-all-drug} of the Supplemental material.

\begin{figure}[t]
\centering
\includegraphics[width=0.42\textwidth]{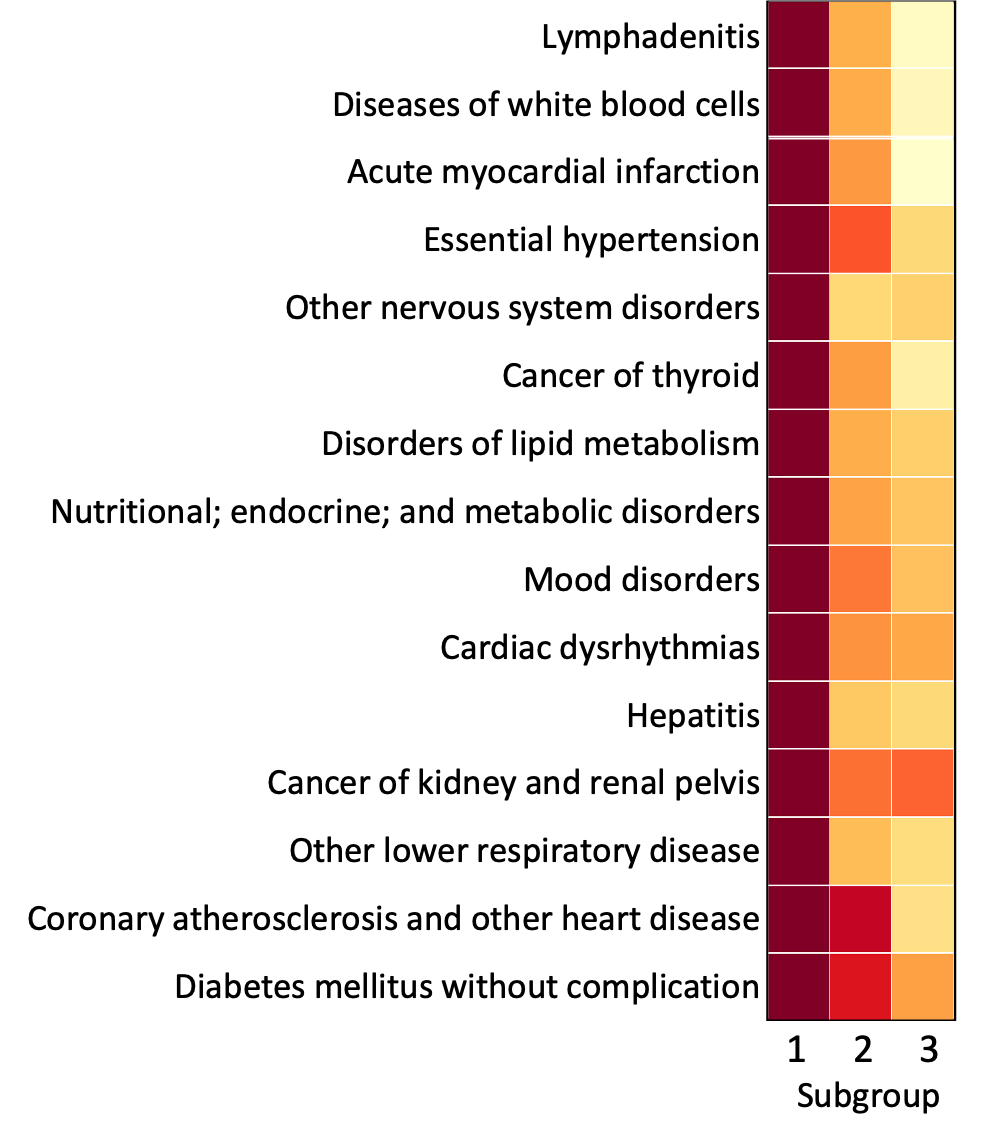}
    \caption{The heatmap of the relative attention scores of covariates among the identified subgroups for Trazodone.} 
    \label{fig-hm}
\end{figure}

\vspace{3pt}
\noindent\textbf{Comparison of Model Conclusion with Supporting Evidence.}
We compare the conclusions from our method with the literature to validate our findings and potential candidates for drug repurposing. Several observational studies and meta-analyses often reported a strong protective effect of statins \cite{haag2009statins, chu2018use, geifman2017evidence, poly2020association, zhang2018statins}. Specifically, Rosuvastatin has shown potential in reducing the risk of progression to dementia in observational studies \cite{xu2023comparing}. However, some randomized controlled trials (RCTs) have reported an absence of positive effects of statins \cite{schultz2018role, sano2011randomized}. The discrepancies between observational studies and RCTs may arise from factors such as patient heterogeneity and variations in health conditions. RCTs generally have strict criteria and controlled conditions. \model addresses these discrepancies using large-scale EHRs, capturing a broader range of responses to statins across diverse populations by considering patient subgroup-specific treatment effects. Notably, our findings show that while Rosuvastatin can be beneficial across all subgroups, Pravastatin appears to be subgroup-specific, suggesting that its effects are not universally applicable. Gabapentin \cite{supasitthumrong2019gabapentin}, Bupropion \cite{das2021drug}, Citalopram \cite{xu2023comparing}, and Trazodone \cite{la2019long} have been shown to have possible benefit or be repurposable in AD patients. However, our method highlights that these drugs may pose risks to certain subgroups, suggesting that they are not beneficial across the entire population. 


\begin{figure}[t]
\centering
\includegraphics[width=0.36\textwidth]{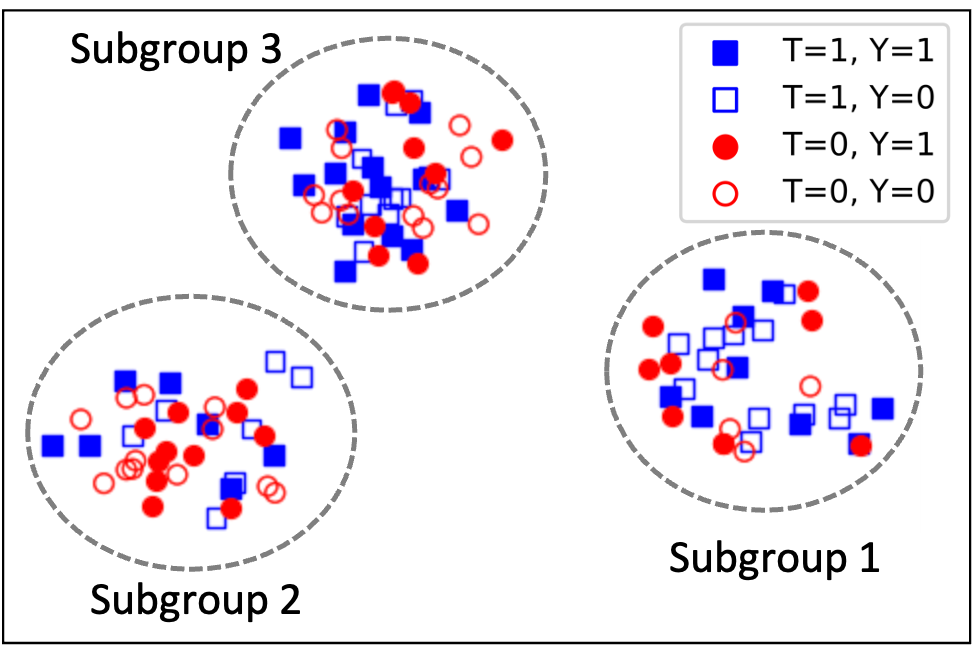}
\vspace{-5pt}
    \caption{Projection scatter plot of the local features for patient subgroups extracted by \model for Trazodone, categorized by their treatment assignments and outcomes. $T=1$ indicates a treated patient, and $Y=1$ represents an adverse outcome.}\label{fig-projection}
\vspace{-10pt}
\end{figure}

\vspace{3pt}
\noindent\textbf{Interpretation of the Enhanced Subgroup.}
We analyze the key characteristics of the subgroup with the most enhanced effects, whose patients may potentially benefit from the trial drug, to further demonstrate the interpretability of our framework. To exemplify this, we select one case drug candidate, Trazodone. We analyze the attention scores derived from both covariate-level and visit-level attention, to investigate the impacts of various covariates on subgroup assignment. The attention scores are averaged within each subgroup and then normalized across all subgroups. These relative attention scores are calculated using the formula: $\Tilde{a}_{i,k}/\sum^{K}_{k=1} \Tilde{a}_{i,k}$, where $\Tilde{a}_{i,k}$ represents the average score of the $i$-th covariate for the $k$-th subgroup patients. Figure \ref{fig-hm} visualizes a heatmap showing the relative scores, with warmer colors (e.g. red) indicating higher scores. We report 15 of covariates with the highest scores on Subgroup 1 (most enhanced group), which are also larger than $1/3$. For example, conditions like lymphadenitis, diseases of white blood cells, and acute myocardial infarction are given more importance in this subgroup. The conditions, reported covariates, likely have a more significant association with the subgroup's response to Trazodone. Additionally, as the subgroup's response diminishes, from Subgroup 1 to Subgroup 3, the scores also decrease progressively. This trend suggests that the patients with these conditions are more responsive to Trazodone. The analysis provides an understanding of the enhanced subgroup and explains why the drug may be more effective for this subgroup. Such findings are instrumental in guiding more personalized drug strategies, aligning with the goal of our research to enhance the precision of drugs. 

Figure \ref{fig-projection} visualizes the local features of patient subgroups extracted by \model for Trazodone, categorized by their treatment assignments and outcomes, using t-SNE algorithm \citep{van2008visualizing}. We randomly sampled 30 examples for each category to show the treatment and outcome distributions by subgroup and compare them. Notably, the subgroups are distinctly separated, indicating the model captures subgroup-specific differences. Subgroup 1 includes more treated patients with non-adverse outcomes (T=1, Y=0) and control patients with adverse outcomes (T=0, Y=1), indicating a positive response to the treatment. In contrast, Subgroup 3 shows a higher proportion of treated patients with adverse outcomes, suggesting a diminished effect of Trazodone for this group.

\section{Conclusion}\label{sec-6}
In this work, we address the crucial challenges inherent in computational drug repurposing. We introduce a novel framework that seamlessly integrates patient subgroup identification and TEE for precision drug repurposing. The real-world study demonstrates the efficiency of our method in identifying potential drug candidates and enhancing precise treatment selection. Our work represents a useful framework for precision drug repurposing, especially in its application to complex diseases characterized by varied patient responses. 

\section{Acknowledgments}
We acknowledge the funding for the project provided by the National Science Foundation (2145625) and the National Institutes of Health (R01GM141279).

\bibliographystyle{ACM-Reference-Format}
\bibliography{sample-base}

\appendix
\section{Appendix}
\subsection{Causal Assumption}\label{sec-ci}

Our study is based on three standard assumptions in causal inference \cite{lechner2001identification}, which are as follows: (1) Conditional Independence Assumption: The assignment of treatment is independent of the outcome, given the pre-treatment covariates. (2) Common Support Assumption: There is a nonzero probability of the treatment assignment for all samples. (3) Stable Unit Treatment Value Assumption: The observed outcome of each unit remains unaffected by the assignment of treatments to other units. These assumptions are essential in treatment effect estimation as they provide the necessary conditions for unbiased and consistent estimation of causal effects. The assumptions form the basis for our methodology. 

\noindent\textbf{Assumption 1 (Ignorability)} \textit{Given the observed covariates, the treatment assignment is independent of the potential outcome, such that, $Y(T=t) \perp  T|X$, for $t \in \{0, 1\}$.}

\noindent\textbf{Assumption 2
(Positivity/Overlap)} \textit{Given the observed covariates, the treatment assignment is non-deterministic, such that, $0<P(T=t|X)<1$, for $t \in \{0, 1\}$.}

\begin{table}[t]
\centering
\caption{Statistics on the synthetic and semi-synthetic datasets}\label{tb-datastats}
\vspace{-7pt}
\begin{tabular}{ll|ccc}\toprule
Statistics &  & Synthetic A & Synthetic B & IHDP \\ \midrule
No. of samples       & Total   & 1000     & 1000 & 747     \\
& Case    & 500     & 483 & 139     \\
& Control & 500     & 517 & 608     \\ \midrule
\multirow{2}{*}{\begin{tabular}[c]{@{}l@{}}Avg. of \\ outcomes\end{tabular}} & Total   & 3.93   & 5.01  & 3.18    \\
& Case    & 2.34    & 6.99 & 6.47    \\
& Control & 5.48    & 3.15 & 2.42    \\ \midrule
No. of features      &  & 10 &     25  & 25      \\ \bottomrule
\end{tabular}
\end{table}

\subsection{Simulation study}\label{sec-simulation}
Treatment Effect Estimation (TEE) is crucial in drug repurposing, as it predicts how an existing drug might influence outcomes and quantifies the magnitude of the drug effect. The accuracy and reliability of TEE models, therefore, are directly linked to the success of drug repurposing efforts, suggesting a strong potential for examining new uses for existing drugs.

To assess the predictive performance of our framework, we conduct a quantitative analysis of TEE, comparing it against state-of-the-art neural network-based TEE models \citep{pmlr-v130-curth21a, shi2019adapting, shalit2017estimating, Hassanpour2020Learning, schwab2020learning, nie2021vcnet, zhang2022exploring}. Since treated and control outcomes are not simultaneously observable in real-world data, our study employs two synthetic and one semi-synthetic datasets for the quantitative analysis. We also conduct comparative experiments with two representative subgrouping models \citep{lee2020robust, nagpal2020interpretable} to further evaluate the effectiveness of the proposed method in subgroup identification.

\subsubsection{Datasets}
We use two synthetic and one semi-synthetic datasets, which have both simulated treated and control outcomes for quantitative analysis. The synthetic datasets are entirely composed of simulated data, whereas the semi-synthetic dataset combines real covariates with simulated potential outcomes. The statistics of the datasets are presented in Table \ref{tb-datastats}.

\paragraph{\textbf{Synthetic Dataset A}}
We simulate a synthetic dataset, following existing works \cite{lee2020robust, argaw2022identifying}. The dataset is inspired by the initial clinical trial results of remdesivir to COVID-19 \cite{wang2020remdesivir}. The results show that the shorter the time from the onset of symptoms to the start of clinical trials with Remdesivir, the faster the time to clinical improvement. The dataset comprises 10 covariates, each derived from a specific normal distribution. The covariates are: age $\sim \mathcal{N}(66, 4)$, white blood cell count ($\times10^9$ per L) $\sim \mathcal{N}(66, 4)$, lymphocyte count ($\times10^9$ per L) $\sim \mathcal{N}(0.8, 0.1)$, platelet count ($\times10^9$ per L) $\sim \mathcal{N}(183, 20.4)$, serum creatinine (U/L) $\sim \mathcal{N}(68, 6.6)$, aspartate aminotransferase (U/L) $\sim \mathcal{N}(31, 5.1)$, alanine aminotransferase (U/L) $\sim \mathcal{N}(26, 5.1)$, lactate dehydrogenase (U/L) $\sim \mathcal{N}(339, 51)$, creatine kinase (U/L) $\sim \mathcal{N}(76, 21)$, and time from symptom onset to trial start (days) $\sim \text{Unif}(4, 14)$. For treatment and control outcomes, we first employ the 'Response Surface' from \cite{hill2011bayesian} for all covariates apart from time. A logistic function is then applied to the time covariate to reflect quicker clinical improvement with earlier treatment initiation. Control and treated outcomes are generated as follows:
\begin{align}
    Y(T=0) \sim \mathcal{N}\left (\mathbf{X}_{-0}\boldsymbol{\beta} + \left(1 + e^{-(x_0-9)}\right)^{-1} + 5, 0.1\right) \\
    Y(T=1) \sim \mathcal{N}\left (\mathbf{X}_{-0}\boldsymbol{\beta} + 5 \cdot \left(1 + e^{-(x_0-9)} \right)^{-1}, 0.1\right )
\end{align}
where $\mathbf{X}_{-0}$ indicates standardized covariate values excluding the time covariate $x_0$. Coefficients $\boldsymbol{\beta}$ are selected from $(0, 0.1, 0.2, 0.3, 0.4)$ with respective probabilities of $(0.6, 0.1, 0.1, 0.1, 0.1)$. In total, the dataset consists of 1,000 samples, evenly divided into 500 cases and 500 controls.

\paragraph{\textbf{Synthetic Dataset B}} The data is generated using an autoregressive simulation model with 25 covariates. The time-dependent coefficients on five sequences are simulated from normal distributions, where their means are different for each sequence. For each patient, initial covariates are drawn from a normal distribution $(x_0 \sim \mathcal{N}(0, 10))$. Subsequent covariates are generated using weighted sums of the five previous covariates with the corresponding time-dependent coefficients, and Gaussian noise is added to these covariates. Outcomes are calculated based on the final covariate vectors, following the 'Response Surface' from \cite{hill2011bayesian}. The number of sequences (timesteps) is randomly selected from a discrete uniform distribution for each patient within the range $\{10, 20\}$, and the treatment assignment is determined by a binomial distribution. The final dataset includes covariate histories, treatment assignments, and treated and control outcomes.

\paragraph{\textbf{Semi-synthetic Dataset}}
For the semi-synthetic dataset, we utilize the Infant Health and Development Program (IHDP) dataset \cite{gross1993infant}, which was originally collected from a randomized experiment aimed at evaluating the impact of early intervention on reducing developmental and health problems among low birth weight, premature infants. The dataset consists of 608 control patients and 139 treated patients, totaling 747 individuals, with 25 covariates. The outcomes are simulated based on the real covariates using the standard non-linear mean outcomes of 'Response Surface B' \cite{hill2011bayesian}.


%

\begin{table*}[!ht]
\centering
\caption{Comparison of prediction performance on the synthetic and semi-synthetic datasets. The average score and standard deviation under 30 runs are reported.}\label{tb2}
\vspace{-7pt}
\begin{tabular}{lcclcclcc}
\toprule
\multirow{2}{*}{Method} & \multicolumn{2}{c}{Synthetic A} &  & \multicolumn{2}{c}{Synthetic B} &  & \multicolumn{2}{c}{IHDP} \\ \cline{2-3} \cline{5-6} \cline{8-9} 
 & PEHE $\downarrow$ & $\epsilon$ATE $\downarrow$ &  & PEHE $\downarrow$ & $\epsilon$ATE $\downarrow$ &  & PEHE $\downarrow$ & $\epsilon$ATE $\downarrow$ \\ \toprule
TNet \citep{pmlr-v130-curth21a} & $0.090\pm0.011$  & $0.011\pm0.008$  &  & $-$ & $-$ &  & $0.167\pm0.018$  & $0.043\pm0.021$  \\
SNet \citep{pmlr-v130-curth21a} & $0.084\pm0.014$  & $0.059\pm0.024$  &   & $0.057\pm0.001$  & $0.028\pm0.005$  &  & $0.073\pm0.013$  & $0.027\pm0.016$  \\
DragonNet \citep{shi2019adapting} & $0.081\pm0.013 $ & $0.016\pm0.013$ &  & $0.057\pm0.002$ & $0.035\pm0.014$ &  & $0.105\pm0.037$ & $0.040\pm0.010$ \\
TARNet \citep{shalit2017estimating} & $0.068\pm0.010$ & $0.023\pm0.003$ &  & $0.054\pm0.001$ & $0.024\pm0.001 $ &  & $0.092\pm0.019$ & $0.039\pm0.010$ \\
DR-CRF \citep{Hassanpour2020Learning} & $0.079\pm0.012$  & $0.022\pm0.020$  &  & $0.056\pm0.001$  & $0.027\pm0.012$  & & $0.070\pm0.012$  &  $0.021\pm0.012$  \\
DRNet \citep{schwab2020learning} & $0.076\pm0.012$ & $0.024\pm0.002$ &  & $0.054\pm0.001$ & $0.024\pm0.001$ &  & $0.057\pm0.013 $ & $0.026\pm0.011$ \\ 
VCNet \citep{nie2021vcnet} & $0.034\pm0.005$ & $0.018\pm0.010$ &  & $0.056\pm0.003 $ & $0.038\pm0.011 $ &  & $0.080\pm0.031$ & $0.065\pm0.049$ \\
TransTEE \citep{zhang2022exploring} & $0.045\pm0.046$ & $0.026\pm0.055$ &  & $0.057\pm0.003$ & $0.038\pm0.028$ &  & $0.099\pm0.071$ & $0.153\pm0.046 $ \\ \midrule
\textbf{\model} (ours) & $\mathbf{0.022\pm0.003}$ & $\mathbf{0.010\pm0.008}$ &  & $\mathbf{0.051\pm0.002}$ & $\mathbf{0.011\pm0.006}$ &  & $\mathbf{0.037\pm0.004}$ & $\mathbf{0.024\pm0.018}$ \\
\textit{\quad w/o GMM} & $0.031\pm0.002$ & $0.025\pm0.004$ &  & $0.056\pm0.001$ & $0.020\pm0.003$ &  & $0.043\pm0.003$ & $0.026\pm0.012$ \\
\textit{\quad w/o Attention} & $0.029\pm0.003$ & $0.020\pm0.006$ &  & $0.054\pm0.002$ & $0.018\pm0.005$ &  & $0.041\pm0.004$ & $0.028\pm0.0015$ \\ \bottomrule
\end{tabular}
\end{table*}

\begin{table*}[!ht]
\centering
\caption{Comparison of subgrouping performance on the Synthetic A and IHDP datasets. The average score and standard deviation under 30 trials are reported.}\label{tb-sub-app}
\vspace{-7pt}
\begin{tabular}{lcccccccc}
\toprule
\multicolumn{2}{c}{}  & \multicolumn{3}{c}{Synthetic} & \multicolumn{1}{c}{}   & \multicolumn{3}{c}{IHDP}  \\   \cline{3-5} \cline{7-9} 
\multicolumn{1}{l}{Model}     &      & $V_{within}\downarrow$ & $V_{across}\uparrow$ & PEHE $\downarrow$  &  & $V_{within}\downarrow$ & $V_{across}\uparrow$ & PEHE $\downarrow$    \\ \midrule
R2P \citep{lee2020robust}    & & $0.262 \pm 0.05$     & \textbf{$2.323 \pm 0.09$} & $0.078 \pm 0.07$  &   & $0.500 \pm 0.15$     & $0.643 \pm 0.13$     & $0.154 \pm 0.05$     \\
HEMM \citep{nagpal2020interpretable}    & & $0.407 \pm 0.06$     & $\mathbf{2.233 \pm 0.11}$     & $0.121 \pm 0.03$  &  & $0.570 \pm 0.11$     & $0.591 \pm 0.15$     & $0.172 \pm 0.00$     \\ \midrule
 \textbf{\model} (ours) && $\mathbf{0.235 \pm 0.02}$  & $2.327 \pm 0.05$    &  $\mathbf{0.022\pm0.003}$ && $\mathbf{0.486\pm0.05}$ & $\mathbf{0.698\pm0.08}$ & $\mathbf{0.037\pm0.004}$ \\ \bottomrule
\end{tabular}
\end{table*}

\subsubsection{Baselines}
The following is a concise overview of the baseline models for TEE:
\begin{itemize}
    \item \textbf{TNet} \citep{pmlr-v130-curth21a} is a deep neural network version of T-learner \citep{kunzel2019metalearners}.
    \item \textbf{SNet} \citep{pmlr-v130-curth21a} learns disentangled representations of the covariates, assuming that the covariates can be disentangled into five components, and predicts two potential outcomes and the propensity score by using different components.
    \item \textbf{DrangonNet} \citep{shi2019adapting} consists of a shared feature network and three auxiliary networks that predict propensity score, and treated and control outcomes, respectively.
    \item \textbf{TARNet} \citep{shalit2017estimating} consists of a shared feature network for balanced hidden representations and two auxiliary networks that predict treated and control outcomes, respectively.
    \item \textbf{DR-CRF} \citep{Hassanpour2020Learning} predicts the potential outcomes and the propensity score by learning disentangled representations of the covariates.
    \item \textbf{DRNet} \citep{schwab2020learning} consists of shared base layers, intermediary treatment layers, and heads for the multiple treatment settings with an associated dosage parameter. 
    \item \textbf{VCNet} \citep{nie2021vcnet} uses separate prediction heads for treatment mapping to preserve and utilize treatment information.
    \item \textbf{TransTEE} \citep{zhang2022exploring} leverages the Transformer to model the interaction between the input covariate and treatment.
\end{itemize} 

We also conduct comparative experiments with two representative existing subgrouping models to further evaluate the effectiveness of the proposed method in subgroup identification. 
\begin{itemize}
\item \textbf{R2P} \citep{lee2020robust} is a tree-based recursive partitioning method.
\item \textbf{HEMM} \citep{nagpal2020interpretable} utilizes Gaussian mixture distributions to learn subgroup probabilities.
\end{itemize} 

R2P and HEMM use the CMGP \citep{alaa2017bayesian} model and a neural network-based model, respectively, to pre-estimate treatment effects for subgroup identification.

\subsubsection{Evaluation Metrics} 
We employ the precision in estimating heterogeneous effects (PEHE) metric to measure the treatment effect at the individual level, expressed as: 
\begin{equation}
    \textrm{PEHE} = \frac{1}{N}\sum_{i=1}^N (f_{y_1}(\mathbf{x}_i)-f_{y_0}(\mathbf{x}_i)-\mathbb{E}[y_1-y_0|\mathbf{x}_i])^2
\end{equation}

Additionally, we employ the absolute error in average treatment effect ($\epsilon$ATE) to assess the overall treatment effect at the population level, defined as:
\begin{equation}
    \textrm{$\epsilon$ATE} = \left |\mathbb{E}[f_{y_1}(\mathbf{x})-f_{y_0}(\mathbf{x})]-\mathbb{E}[y_1-y_0]  \right |
\end{equation}

To evaluate the performance for subgroup identification, we analyze the variance of treatment effects within and across subgroups. The variance across the subgroups evaluates the variance of the mean of the treatment effects in each subgroup, while the variance within subgroups measures the mean of the variance of the treatment effects in each subgroup. They are expressed as: 

\begin{equation}
    V_{across}= Var\left (\{Mean(TE_k)\}_{k=1}^K\right )
\end{equation}

\begin{equation}
    V_{within} = \frac{1}{K}\sum_{k=1}^KVar(TE_k)
\end{equation}
where $TE_k$ indicates a set of  estimated treatment effects in subgroup $k$, such that $TE_k = \{\mathbb{E}[y|t=1,\mathbf{x}]-\mathbb{E}[y|t=0,\mathbf{x}]$ for all $\mathbf{x}\in C_k\}$.

\subsubsection{Implementation Details} All neural network-based models are implemented using PyTorch. We use the SGD/Adam optimizer, with a learning rate set to $0.001$ and a batch size of 128. The hyper-parameters for the baseline models follow the implementations provided by the respective authors. For our proposed model, we set the hidden nodes from the set \{50, 100, 200, 300\}, the number of hidden layers in the Transformer and the prediction network from the set \{1,2,3\}, the number of subgroups from the range of [2, 7], and coefficient $\alpha$ from the range of [0.1, 0.5]. The data is randomly divided into training, validation, and test sets, with a split ratio of 6:2:2. We train the model on the training set and employ a stopping rule based on the performance on the validation set. All results are reported on the test set.



\subsubsection{Experimental Results}

\paragraph{\textbf{Treatment Effect Estimation}}

Table~\ref{tb2} shows predictive performances, PEHE and $\epsilon$ATE, on three synthetic and semi-synthetic datasets. Our proposed method shows superior performance, outperforming the baseline models. Specifically, our model exhibits PEHE values of $0.022$ and $0.051$ on the synthetic datasets and $0.037$ on the IHDP dataset. These results are not merely numerically superior but represent a significant reduction in error — 35.3\%, 5.6\%, and 35.1\% lower than the second-best model. This enhancement underscores the efficiency of our approach, which estimates the subgroup-specific treatment effect by identifying heterogeneous subgroups. We also conduct an ablation study to assess the contribution of specific components to the overall performance. This includes two variations: (1) \textit{w/o GMM}, \model without the mixture Gaussian model of the local distribution (i.e., without subgrouping); (2) \textit{w/o Attention}, \model without both covariate-level and visit-level attention mechanisms (without the patient-level representation). The results indicate that each component plays an important role as the performance decreases when removed.

\begin{figure*}[!t]
\centering
\begin{subfigure}[b]{0.25\textwidth}
    \centering
\includegraphics[width=\textwidth]{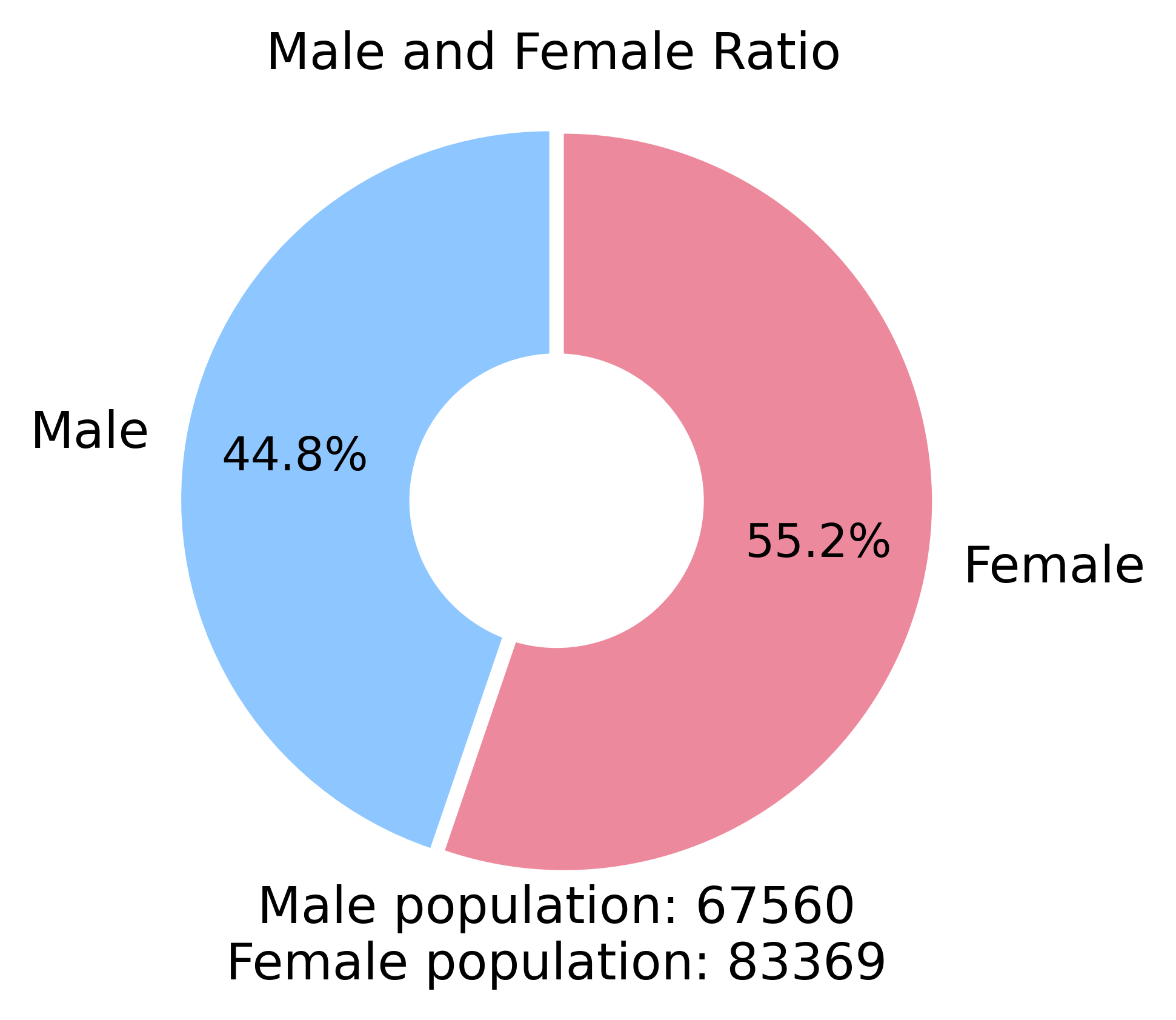}
\end{subfigure}
\begin{subfigure}[b]{0.25\textwidth}
    \centering
    \includegraphics[width=\textwidth]{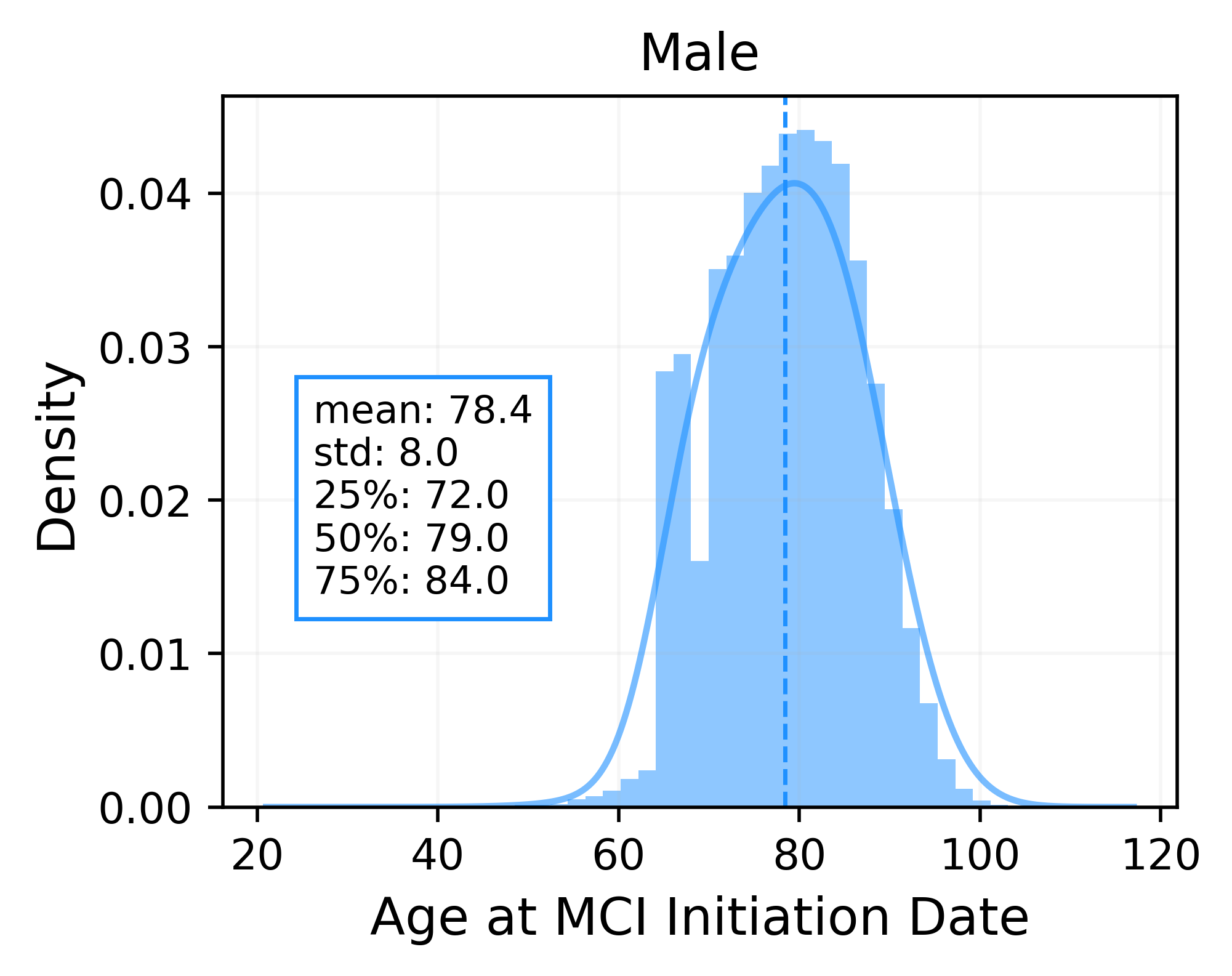}
\end{subfigure}
\begin{subfigure}[b]{0.25\textwidth}
    \centering
    \includegraphics[width=\textwidth]{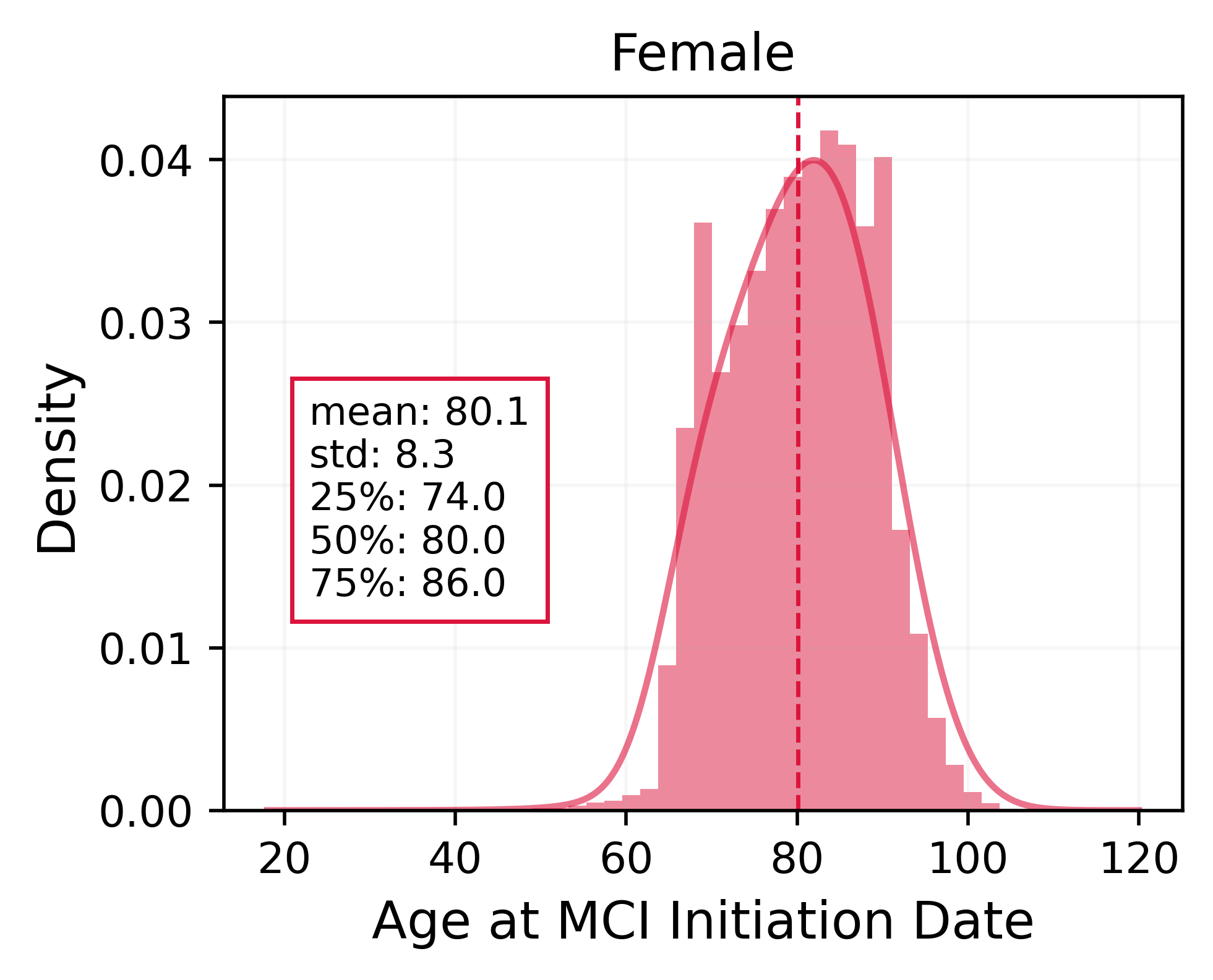}
\end{subfigure}
\vspace{-7pt}
\caption{The distribution of gender and age at MCI initiation date.}\label{fig-age}
\end{figure*}

\paragraph{\textbf{Subgroup Identification}}
A critical dimension of our research is the identification of subgroups. The ideal subgrouping strategy would maximize heterogeneity across different subgroups while ensuring homogeneity within each subgroup. Table \ref{tb-sub-app} presents the subgrouping performance, $V_{within}$ and $V_{across}$, as well as PEHE on the Synthetic A and IHDP datasets. The Synthetic Dataset B is not used since it is not applicable for the baselines (timeseries data). Our method outperforms the baselines on both the subgrouping and TEE metrics across all datasets, except for $V_{across}$ on the Synthetic dataset. Specifically, the baseline models exhibit lower performance in PEHE. This suggests that the baseline models, which rely on fixed pre-trained TEE models to identify subgroups, may not be optimal for this task. Consequently, the subgroup performance metrics of these models are also shown to be lower. This observation highlights a fundamental limitation in existing subgrouping methods. In contrast, our approach improves performance in both subgroup identification and TEE by simultaneously optimizing subgrouping and TEE. 



\begin{table*}[t]
\centering
\caption{The definition of cognitive impairment (MCI) and Alzheimer’s Disease (AD).}\label{tb-ad}
\vspace{-7pt}
\begin{tabular}{l|l}
\toprule
MCI& \begin{tabular}[c]{@{}l@{}}
\textbf{ICD-9}:\\ 331.83 Mild cognitive impairment, so stated\\ 294.9 Unspecified persistent mental disorders due to conditions classified elsewhere\\
\textbf{ICD-10}:\\ G31.84 Mild cognitive impairment, so stated\\ F09 Unspecified mental disorder due to known physiological condition\end{tabular}\\ \midrule
AD& \begin{tabular}[c]{@{}l@{}}
\textbf{ICD-9}:\\ 331.0 Alzheimer’s disease\\ 
\textbf{ICD-10}:\\ G30.* Alzheimer’s disease\end{tabular}\\ \midrule
\begin{tabular}[c]{@{}l@{}}ADRD\end{tabular} & \begin{tabular}[c]{@{}l@{}}
\textbf{ICD-9}:\\ 294.10 Dementia in conditions classified elsewhere without behavioral disturbance\\ 294.11 Dementia in conditions classified elsewhere with behavioral disturbance\\ 294.20 Dementia, unspecified, without behavioral disturbance.\\ 294.21 Dementia, unspecified, with behavioral disturbance\\ 290.* Dementias\\
\textbf{ICD-10}:\\ F01.* Vascular dementia\\ F02.* Dementia in other diseases classified elsewhere\\ F03.* Unspecified dementia\end{tabular} \\ \bottomrule
\end{tabular}
\end{table*}

\subsection{Drug Repurposing for Alzheimer’s Disease}\label{sec-AD}

\subsubsection{Dataset}
We used a large-scale real-world longitudinal patient-level healthcare warehouse, MarketScan\footnote{\url{https://www.merative.com/real-world-evidence}} Medicare Supplemental and Coordination of Benefits Database (MDCR) from 2012 to 2018, which contains individual-level and de-identified healthcare claims information. The MarketScan MDCR database is created for Medicare-eligible retirees with employer-sponsored Medicare Supplemental plans. The MarketScan data are primarily used to evaluate health utilization and services, containing administrative, including patients' longitudinal information, including diagnoses, procedures, prescriptions, and demographic characteristics. We identify 155K distinct MCI patients, among whom 40K are diagnosed with AD. The distribution of gender and age at MCI initiation date and patients’ distribution of total time in the database are shown in Figure \ref{fig-age}. MCI and AD patients are identified using the diagnosis codes, as shown in Table \ref{tb-ad}. We consider diagnosis and prescription codes for study variables. The diagnosis codes are defined by the International Classification of Diseases (ICD) 9/10 codes. We map the ICD codes to Clinical Classifications Software (CCS), including a total of 286 codes. For the drugs, we match national drug codes (NDCs) to ingredient levels, resulting in a total of 1,353 unique drugs in our dataset. We use diagnosis codes and their time information to construct input variables for our method.

\subsubsection{Experimental Setup}
In our study, we set the number of clusters to three for all trial drugs. For each drug trial emulation, the data is randomly divided into training, validation, and test sets, with a split ratio of 6:2:2. We train the model on the training set and employ a stopping rule based on the performance on the validation set. The hyperparameters of our model include $\alpha$ which is a penalty strength factor for $\mathcal{L}_{overlap}$, the number of layers and hidden nodes in the Transformer encoder, the number of hidden nodes in both the local/global distribution in the subgroup representation network, and the number of layers of the prediction network. Notably, the number of hidden nodes in the prediction network is the same as that in the subgroup representation network. We evaluated our model across several trials to determine the optimal hyperparameters and applied these parameters consistently across all drug evaluations. Detailed information about the hyperparameters can be found on our GitHub repository\footnote{\url{https://github.com/yeon-lab/STEDR}}.

\subsubsection{Metrics}
We measure the balance between case and control cohorts by standardized mean difference (SMD) \cite{austin2009using} and weighted propensity score area under the curve (AUC) \cite{shimoni2019evaluation} using IPTW from our model to evaluate selection bias. For a continuous covariate, the SMD is calculated as follows:
\begin{equation}
     SMD = \frac{(\bar{x}_{treatment}-\bar{x}_{control})}{\sqrt{\frac{s^2_{treatment}+s^2_{control}}{2}}}
\end{equation}
where $\bar{x}_{treatment}$ and $\bar{x}_{control}$ represent the mean of the covariate in patients from case and control cohorts, respectively, while $s^2_{treatment}$ and $s^2_{control}$ denote the variance of the covariate. The covariate is considered balanced if its $SMD \leq 0.1$ \cite{zang2022high, austin2009using}. Emulated trials are regarded as balanced if the proportion of unbalanced covariates is $\leq 2\%$ of all covariates \cite{liu2021deep, zang2022high}.

\subsubsection{Experimental Results}

\begin{figure*}[!hb]
    \centering
\includegraphics[width=0.85\textwidth]{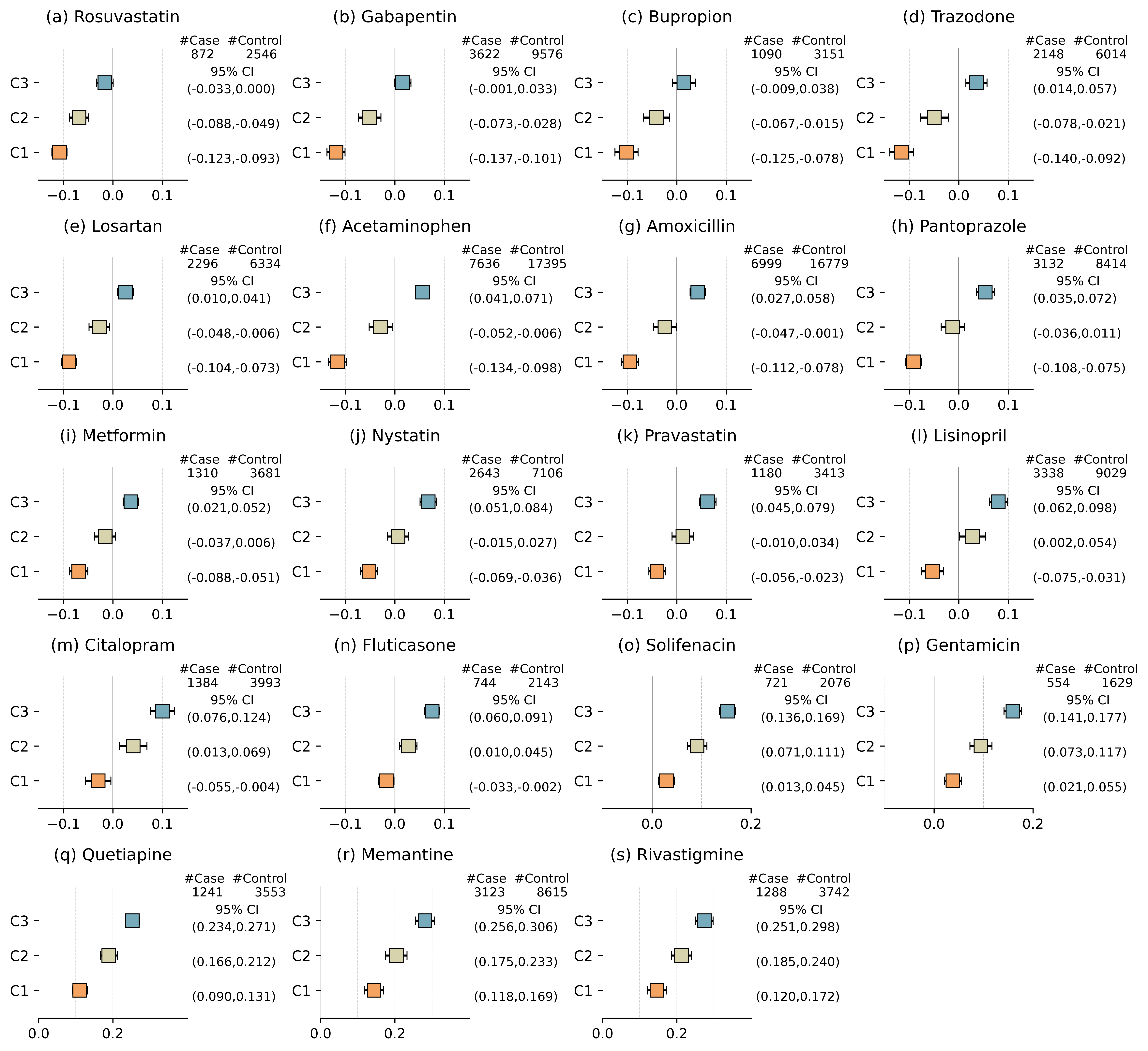}
\caption{Visualization of 95\% confidence intervals of estimated treatment effects across different patient subgroups for 19 trial drugs. C1 to C3 represents Subgroups 1 to 3, with Subgroup 1 showing the most enhanced effect and Subgroup 3 exhibiting the most diminished effect.}\label{fig-all-drug}
\end{figure*}

\begin{table*}[!hb]
\caption{Estimated treatment effects with 95\% CIs for subgroups of trial drugs. The number of unbalanced features is computed after IPTW.}\label{tb-all-drug}
\begin{tabular}{lllllccc}
\toprule
\multirow{3}{*}{Trial Drug} & \multirow{3}{*}{\begin{tabular}[c]{@{}l@{}}No. of \\ Cases\end{tabular}} & \multirow{3}{*}{\begin{tabular}[c]{@{}l@{}}No. of\\ Controls\end{tabular}} & \multirow{3}{*}{\begin{tabular}[c]{@{}l@{}}No. of \\ unbalanced \\ feat.\end{tabular}} & \multirow{3}{*}{$p$ value} & \multicolumn{3}{c}{Estimated Effects (Mean [CI])} \\ \cline{6-8} 
 &  &  &  &  & \multirow{2}{*}{Subgroup 1} & \multirow{2}{*}{Subgroup 2} & \multirow{2}{*}{Subgroup 3} \\
 &  &  &  &  &  &  &  \\ \hline
Rosuvastatin & 872 & 2546 & 3.6 & \textless .001 & -0.11 (-0.12,-0.09) & -0.07 (-0.09,-0.05) & -0.02 (-0.03, 0.00) \\ \midrule
Gabapentin & 3622 & 9576 & 4.9 & \textless .001 & -0.12 (-0.14,-0.10) & -0.05 (-0.07,-0.03) & 0.02 (-0.00, 0.03) \\ \midrule
Trazodone & 2148 & 6014 & 1.7 & \textless .001 & -0.12 (-0.14,-0.09) & -0.05 (-0.08,-0.02) & 0.04 ( 0.01, 0.06) \\ \midrule
Bupropion & 1090 & 3151 & 3.8 & \textless .001 & -0.10 (-0.13,-0.08) & -0.04 (-0.07,-0.01) & 0.01 (-0.01, 0.04) \\ \midrule
Losartan & 2296 & 6334 & 3.8 & \textless .001 & -0.09 (-0.10,-0.07) & -0.03 (-0.05,-0.01) & 0.03 ( 0.01, 0.04) \\ \midrule
Acetaminophen & 7636 & 17395 & 2.5 & \textless .001 & -0.12 (-0.13,-0.10) & -0.03 (-0.05,-0.01) & 0.06 ( 0.04, 0.07) \\ \midrule
Amoxicillin & 6999 & 16779 & 4.2 & \textless .001 & -0.09 (-0.11,-0.08) & -0.02 (-0.05,-0.00) & 0.04 ( 0.03, 0.06) \\ \midrule
Metformin & 1310 & 3681 & 3.1 & \textless .001 & -0.07 (-0.09,-0.05) & -0.02 (-0.04, 0.01) & 0.04 ( 0.02, 0.05) \\ \midrule
Pantoprazole & 3132 & 8414 & 2.9 & \textless .001 & -0.09 (-0.11,-0.08) & -0.01 (-0.04, 0.01) & 0.05 ( 0.04, 0.07) \\ \midrule
Nystatin & 2643 & 7106 & 2.2 & \textless .001 & -0.05 (-0.07,-0.04) & 0.01 (-0.01, 0.03) & 0.07 ( 0.05, 0.08) \\ \midrule
Pravastatin & 1180 & 3413 & 1.7 & \textless .001 & -0.04 (-0.06,-0.02) & 0.01 (-0.01, 0.03) & 0.06 ( 0.05, 0.08) \\ \midrule
Fluticasone & 744 & 2143 & 3.8 & 0.029 & -0.02 (-0.03,-0.00) & 0.03 ( 0.01, 0.04) & 0.08 ( 0.06, 0.09) \\ \midrule
Lisinopril & 3338 & 9029 & 4.8 & \textless .001 & -0.05 (-0.08,-0.03) & 0.03 ( 0.00, 0.05) & 0.08 ( 0.06, 0.10) \\ \midrule
Citalopram & 1384 & 3993 & 1.4 & 0.024 & -0.03 (-0.06,-0.00) & 0.04 ( 0.01, 0.07) & 0.10 ( 0.08, 0.12) \\ \midrule
Solifenacin & 721 & 2076 & 3.6 & \textless .001 & 0.03 ( 0.01, 0.05) & 0.09 ( 0.07, 0.11) & 0.15 ( 0.14, 0.17) \\ \midrule
Gentamicin & 554 & 1629 & 3.6 & \textless .001 & 0.04 ( 0.02, 0.06) & 0.09 ( 0.07, 0.12) & 0.16 ( 0.14, 0.18) \\ \midrule
Quetiapine & 1241 & 3553 & 3.4 & \textless .001 & 0.11 ( 0.09, 0.13) & 0.19 ( 0.17, 0.21) & 0.25 ( 0.23, 0.27) \\ \midrule
Memantine & 3123 & 8615 & 1.9 & \textless .001 & 0.14 ( 0.12, 0.17) & 0.20 ( 0.18, 0.23) & 0.28 ( 0.26, 0.31) \\ \midrule
Rivastigmine & 1288 & 3742 & 5.1 & \textless .001 & 0.15 ( 0.12, 0.17) & 0.21 ( 0.19, 0.24) & 0.27 ( 0.25, 0.30) \\ \bottomrule
\end{tabular}
\end{table*}





\end{document}